\documentclass{article}

\usepackage{graphicx}
\usepackage{latexsym}

\newtheorem{Theorem}{{\bf Theorem}}
\newtheorem{Lemma}{{\bf Lemma}}
\newtheorem{Corollary}{{\bf Corollary}}

\title{
Equations of States 
in Singular Statistical Estimation}
\author{
Sumio Watanabe\\
Precision and Intelligence Laboratory\\
Tokyo Institute of Technology\\
4259 Nagatsuta, Midori-ku, Yokohama, 226-8503 Japan\\
E-mail: swatanab@pi.titech.ac.jp
}

\begin{document}

\large

\maketitle

\setlength{\baselineskip}{20pt}

\begin{abstract}
Learning machines that have 
hierarchical structures or hidden variables 
are singular statistical models
because they are nonidentifiable and 
their Fisher information matrices are singular.
In singular statistical models, neither 
does the Bayes
{\it a posteriori} distribution converge to the normal distribution
nor does the maximum likelihood
estimator satisfy asymptotic normality.
This is the main reason that it has been difficult to 
predict their generalization performance from trained states. 
In this paper, we study four errors, 
(1) the Bayes generalization error,
(2) the Bayes training error, (3) the Gibbs generalization error, and
(4) the Gibbs training error, and prove that there are universal
mathematical relations among 
these errors. The formulas proved in this paper are equations of states in
statistical estimation because they hold for any true distribution,
any parametric model, and any {\it a priori} distribution. 
Also we show that the Bayes and Gibbs generalization errors can be 
estimated by Bayes and Gibbs training errors, and we propose 
widely applicable information criteria that can be applied to
both regular and singular statistical models. 
\end{abstract}

\section{Introduction}

Recently, many learning machines are being
used in information processing systems. For 
example, layered neural networks, normal mixtures,
binomial mixtures, Bayes networks, Boltzmann machines,
reduced rank regressions, hidden Markov models, and
stochastic context-free grammars 
are being employed in pattern recognition, time series 
prediction, robotic control, human modeling, 
and biostatistics. 
Although their generalization performances determine
the accuracy of the information systems, 
it has been difficult to estimate generalization 
errors based on training errors, because such 
learning machines are singular statistical models. 

A parametric model is called regular
if the mapping from the parameter to the 
probability distribution is one-to-one and if
its Fisher information matrix is always positive definite. 
If a statistical model is regular, then 
the Bayes {\it a posteriori}
distribution converges to the normal distribution, 
and the maximum likelihood estimator satisfies 
asymptotic normality. Based on such properties, 
the relation between
the generalization error and the training error
was clarified, on which some information criteria 
were proposed. 

On the other hand, if the mapping from the 
parameter to the probability distribution is not
one-to-one or if the
Fisher information matrix is 
singular, then the parametric model is called singular.
In general, if a learning machine has hierarchical structure
or hidden variables, then it is singular. Therefore,
almost all learning machines are singular. 
For singular learning machines, the log likelihood 
function can not be approximated by any
quadratic form of the parameter, with the result that
the conventional
relationship between generalization errors and 
training errors does not hold either for
the maximum likelihood method \cite{Hartigan}
\cite{Hagiwara}\cite{Hayasaka} or 
Bayes estimation \cite{1995}. Singularities
strongly affect generalization performances
\cite{2001a} and learning dynamics \cite{Amari}.
Therefore, in order
to establish the mathematical foundation of
singular learning theory, it is necessary 
to construct the formulas which hold
even in singular learning machines. 

Recently, we proved \cite{1999}\cite{2001a} that the 
generalization error in Bayes estimation
is asymptotically equal to $\lambda/n$,
where $\lambda>0$ is the rational number 
determined by the zeta function of a learning machine
and $n$ is the number of training samples. 
In regular statistical models, $\lambda=d/2$, 
where $d$ is the dimension of the parameter space,
whereas in singular statistical models, $\lambda$ 
depends strongly on the learning machine,
the true distribution, and the {\it a priori}
probability distribution. 
In practical applications, the true distribution
is often unknown, hence it has been difficult to
estimate the generalization error from the
training error. To estimate the generalization error
when we do not have any information about the
true distribution, we need a general formula which
holds independently of singularities. 

In this paper, we study four errors, 
(1) the Bayes generalization error $B_{g}$,
(2) the Bayes training error $B_{t}$, (3) the Gibbs generalization error $G_{g}$,
and (4) the Gibbs training error $G_{t}$, and prove the formulas
\begin{eqnarray*}
E[B_{g}]-E[B_{t}]&=&2\beta(E[G_{t}]-E[B_{t}])+o(\frac{1}{n}),\\
E[G_{g}]-E[G_{t}]&=&2\beta(E[G_{t}]-E[B_{t}])+o(\frac{1}{n}),
\end{eqnarray*}
where $E[\cdot]$ denotes the expectation value and 
$0<\beta<\infty$ is the inverse temperature 
of the {\it a posteriori} distribution. These equations
assert that the increased error from training to 
generalization is in proportion to 
the difference between the Bayes and Gibbs training errors. 
It should be emphasized that these formulas
hold for any true distribution, any learning machine, 
any {\it a priori} probability distribution, and any singularities,
therefore they reflect the universal laws of statistical estimation.
Also, 
based on the formula, we propose widely applicable information 
criteria (WAIC) which can be applied to both regular and singular
learning machines. In other words, we can apply WAIC without 
any knowledge about the true distribution. 

This paper consists of six parts. In Section 2, we 
describe the main results of this paper. In Section 3, 
we propose widely applicable information criteria 
and show how to apply them to statistical estimation. 
In Section 4, we prove the main results in the mathematically
rigorous way. In Sections 5 and 6, we discuss and conclude of
this paper. 
The proofs of lemmas are quite technical hence they are
presented in Appendix.

\section{Main Results}

Let $(\Omega,{\cal B},P)$ be a probability space, and 
$X:\Omega\rightarrow {\bf R}^{N}$ be a random variable whose
probability distribution is $q(x)dx$. Here ${\bf R}^{N}$ denotes the
$N$ dimensional Euclidean space. We assume that the random variables 
$X_{1},X_{2},..,X_{n}$ are independently subject to the same probability 
distribution as $X$. In learning theory, 
$q(x)dx$ is called the true distribution and
$
D_{n}=\{X_{1},X_{2},...,X_{n}\}
$
is a set of training samples. A learning machine 
is defined by a parametric probability density function 
$p(x|w)$ of $x\in{\bf R}^{N}$ for a
given parameter $w\in W \subset {\bf R}^{d}$,
where $W$ is a set of parameters. 
An {\it a priori } probability density function $\varphi(w)$ 
is defined on $W$. 
The Bayes {{\it a posteriori} probability density $p(w|D_{n})$ for
a given set of training samples $D_{n}$ is defined by
\[
p(w|D_{n})=\frac{1}{C_{n}}\;\varphi(w)\;
\Bigl{(}
\prod_{i=1}^{n}p(X_{i}|w)
\Bigr{)}^{\beta},
\]
where $\beta>0$ is the inverse temperature and
$C_{n}>0$ is the normalizing constant. 
The expectation
value with respect to this probability distribution is denoted by
$E_{w}[\cdot]$. 
Also $E_{D_{n}}[\cdot]$ and $E_{X}[\cdot]$ denote
respectively the expectation values over $D_{n}$ and
$X$. We sometimes omit $D_{n}$ and simply use $E[\cdot]$. 
We study the four errors, defined below. \\
(1) Bayes generalization error,
\[
B_{g} =
E_{X}\Bigl{[}\log\frac{q(X)}{E_{w}[p(X|w)]}
\Bigr{]}.
\]
(2) Bayes training error,
\[
B_{t}=
\frac{1}{n}\sum_{j=1}^{n}
\log\frac{q(X_{j})}{E_{w}[p(X_{j}|w)]}.
\]
(3) Gibbs generalization error,
\[
G_{g}=E_{w}\Bigl{[}
E_{X} [\log \frac{q(X)}{p(X|w)}]
\Bigr{]}.
\]
(4) Gibbs training error,
\[
G_{t}= E_{w}\Bigl{[}
\frac{1}{n}\sum_{j=1}^{n}
\log\frac{q(X_{j})}{p(X_{j}|w)}
\Bigr{]}.
\]
These four errors are measurable functions of $D_{n}$, 
hence they are also random variables.
\vskip3mm\noindent
{\bf Remark.} The Bayes generalization error is equal to
the Kullback-Leibler distance from the true distribution $q(x)$ to
the Bayes predictive distribution $E_{w}[p(x|w)]$. 
The Gibbs generalization error
is equal to the average of the Kullback-Leibler distance from 
the true distribution to the Gibbs estimation. They show the
accuracy of Bayes and Gibbs estimations, it is important
for statistical learning machines to be able to estimate them from
random samples. 
\vskip3mm
We need some mathematical assumptions which ensure that the theorems hold.
Let us define a log density ratio function by
\[
f(x,w)=\log\frac{q(x)}{p(x|w)}.
\]
In this paper, we mainly study the singular case, that is to say, the situation
when
the set of true parameters $\{w\in W;q(x)=p(x|w)\}$ consists of more than one point and
the Fisher information matrix is not positive definite. 
We assume the following three conditions.\\
{\bf (A.1)} Assume that the set of parameters $W$ is 
a compact set which is the closure of an open set 
in ${\bf R}^{d}$. The set $W$ is defined by 
\[
W=\{w\in{\bf R}^{d};\pi_{1}(w)\geq 0,\cdots \pi_{k}(w)\geq 0\},
\]
where $\pi_{1}(w),\cdots,\pi_{k}(w)$ are analytic functions, 
and the {\it a priori} probability density $\varphi(w)$ is given by
$\varphi(w)=\varphi_{0}(w)\varphi_{1}(w)$ where $\varphi_{0}(w)> 0$ 
is a $C^{\infty}$-class
function and $\varphi_{1}(w)\geq 0$ is an analytic function. 
\\
{\bf (A.2)} Let $s\geq 6$ be a constant, and 
$L^{s}(q)$ be the complex Banach space defined by
\[
L^{s}(q)=\{f(x)\;;\;\int |f(x)|^{s}q(x)dx<\infty \}.
\]
Assume that there exists an open set $W'\subset {\bf C}^{d}$ 
which contains $W$ such that
the function $W'\ni w\mapsto f(\cdot,w)$ is an $L^{s}(q)$
valued analytic function.\\
{\bf (A.3)} Let $W_{0}=\{w\in W\;;\;q(x)=p(x|w)\}$ be the
set of true parameters. The set $W_{0}$ is not the empty set and 
there exists an open set $W^{*}\subset {\bf C}^{d}$ which contains
$W$  such that for $M(x)\equiv \sup_{w\in W{*}}|f(X,w)|$, 
\[
E_{X}[\sup_{w\in W{*}}|f(X,w)|^{s}]<\infty.
\]
and there exists $t>0$ such that, for $\displaystyle Q(x)\equiv \sup_{K(w)\leq t} p(x|w) $
\[
\int M(x)^{2}Q(x)dx<\infty.
\]
\vskip3mm
\noindent{\bf Remark.} These assumptions 
are needed for the mathematical reasons. \\
(1) These conditions
allow for the case that the set of 
true parameters $W_{0}=\{w\in W; q(x)=p(x|w)\}$ 
is not a single point but an algebraic set or an 
analytic set with singularities. In general, 
the Fisher information matrix has zero 
eigenvalues. On the other hand, in conventional statistical
learning theory, it is assumed that $W_{0}$ consists of one point and
the Fisher information matrix is positive definite. 
On the assumptions of this paper, we can not use any result of 
conventional statistical learning theory.
\\
(2) The condition that $W$ is compact is necessary
because, even if the log density ratio function is
an analytic function of the parameter, 
$|w|=\infty$ is a singularity in general. For this reason, 
if $W$ is not compact and $W_{0}$ contains $|w|=\infty$, 
the maximum likelihood estimator
does not exist in general. In fact, if $x=(x_{1},x_{2})$,
$w=(a,b)$, and $f(x,w)=(x_{2}-a\sin(bx_{1}))^{2}/2$, 
and $W_{0}$ contains $\{a=0\}$, 
then the maximum likelihood estimator never exists. 
On the other hand, if $|w|=\infty$ is not a singularity, 
${\bf R}^{d}\cup\{|w|=\infty\}$ can be understood as a 
compact set and the same theorems established in this paper hold. 
 \\
(3) The condition that $\pi_{1}(w),...,\pi_{k}(w)$ and
$\varphi_{1}(w)$ are analytic functions is necessary
because if one of them is a $C^{\infty}$ class
function, there exists a pathological example. In fact,
if $\varphi_{1}(w)=\exp(-1/\|w\|^{2})$ in
a neighborhood of the origin and the set of
true parameters is the origin, then the four errors
may not be in proportion to $1/n$. \\
(4) The condition $s\geq 6$ is needed to ensure
the existence of the asymptotic expansion of the Bayes generalization
error in our proof. (See the proof of Theorem \ref{Theorem:111}.)\\
(5) Some non-analytic statistical models can be made analytic. 
For example, in a simple mixture model
$p(x|a) = a p_{1}(x)+(1-a)p_{2}(x)$ for some
probability densities $p_{1}(x)$ and $p_{2}(x)$, 
the log density ratio function $f(x,a)$ is not analytic at $a=0$, but
it can be made analytic by the representation 
$p(x|\theta)=\alpha^{2} p_{1}(x)+ \beta^{2} p_{2}(x)$,
on the manifold $\theta\in\{\alpha^{2}+\beta^{2}=1\}$. As is shown in the
proofs, if $W$ is contained in an analytic manifold, then the 
same theorems hold as stated in this paper.  \\
(6) Note that 
\begin{equation}
\int M(x)^{6}q(x)dx<\infty.\label{eq:M(x)}
\end{equation}
\vskip3mm\noindent
Based on assumptions (A.1), (A.2), and (A.3), we prove
the following results. 
\begin{Theorem}\label{Theorem:111}
(1) There exist random variables $B^{*}_{g}$,
$B^{*}_{t}$, $G^{*}_{g}$, and $G^{*}_{t}$ such that,
as $n\rightarrow\infty$, 
the following convergences in law hold.
\[
nB_{g} \rightarrow  B^{*}_{g}, \;\;\;
nB_{t} \rightarrow  B^{*}_{t}, \;\;\;
nG_{g} \rightarrow  G^{*}_{g}, \;\;\;
nG_{t} \rightarrow  G^{*}_{t}. 
\]
(2) As $n\rightarrow\infty$, the following convergence in probability holds,
\[
n (B_{g}-B_{t}-G_{g}+G_{t})\rightarrow 0.
\]
(3) The expectation values of the four errors converge as follows, 
\begin{eqnarray*}
E[nB_{g}] \rightarrow  E[B^{*}_{g}], &&
E[nB_{t}] \rightarrow  E[B^{*}_{t}], \\
E[nG_{g}] \rightarrow  E[G^{*}_{g}], &&
E[nG_{t}] \rightarrow  E[G^{*}_{t}]. 
\end{eqnarray*}
\end{Theorem}
For the proof of this theorem, see Section \ref{Section:proof}.
the following Theorem is the main result of this paper. 

\begin{Theorem}\label{Theorem:222} ({\bf Equations of States in Statistical Estimation}).
The following equations hold.
\begin{eqnarray}
E[B^{*}_{g}]-E[B^{*}_{t}]&=& 2\beta(E[G^{*}_{t}]-E[B^{*}_{t}]),\label{eq:ES1}\\
E[G^{*}_{g}]-E[G^{*}_{t}]&=& 2\beta(E[G^{*}_{t}]-E[B^{*}_{t}]).
\label{eq:ES2}
\end{eqnarray}
\end{Theorem}
\vskip3mm\noindent
{\bf Remark}. (1) Theorem \ref{Theorem:222} asserts that 
the increases of errors from training to prediction 
are in proportion to 
the difference between the Bayes and Gibbs training errors. We refer to 
Theorem \ref{Theorem:222} as {\bf Equations of States in Statistical Estimation},
because they hold for any true distribution,
any learning machine, any {\it a priori}
distribution, and any singularities. It is proved that the equations of states
hold even if the true distribution is not contained in the parametric model \cite{NC2009-3}. \\
(2) Although the equations of states hold universally, 
the four errors themselves depend strongly on a
true distribution, a learning machine, an {\it a priori}
distribution, and singularities. 
\\
(3) Theorem \ref{Theorem:222} also asserts a conservation law, namely,
the difference between the Bayes error and the Gibbs error
is invariant between training and generalization, 
\begin{equation}
E[G^{*}_{g}]-E[B^{*}_{g}]=
E[G^{*}_{t}]-E[B^{*}_{t}].
\label{eq:conservation}
\end{equation}
As is shown in Theorem \ref{Theorem:111}, this
conservation law holds not only for expectations, but
also for the random variables, as the number of 
training samples tends to infinity. 
\begin{Corollary}\label{Corollary:222}
The two generalization errors can be
estimated by the two training errors,
\begin{equation}
\left(
\begin{array}{c}
E[B^{*}_{g}] \\
E[G^{*}_{g}]
\end{array}
\right)
=
\left(
\begin{array}{cc}
1-2\beta & 2\beta \\
-2\beta & 1+2\beta
\end{array}
\right)
\left(
\begin{array}{c}
E[B^{*}_{t}] \\
E[G^{*}_{t}]
\end{array}
\right).\label{eq:invariant}
\end{equation}
\end{Corollary}
\vskip3mm\noindent
{\bf Remark.} (1) From eq.(\ref{eq:invariant}), 
it follows that
\[
\left(
\begin{array}{c}
E[G^{*}_{t}] \\
E[B^{*}_{t}]
\end{array}
\right)
=
\left(
\begin{array}{cc}
1-2\beta & 2\beta \\
-2\beta & 1+2\beta
\end{array}
\right)
\left(
\begin{array}{c}
E[G^{*}_{g}] \\
E[B^{*}_{g}]
\end{array}
\right),
\]
which shows that there is a symmetry between generalization errors
and training errors. \\
(2) Since the set of eigenvalues of the linear transform in eq.(\ref{eq:invariant}) 
is $\{1\}$, and the dimension of the linear invariant subspace is one,
there is no conservation law other than eq.(\ref{eq:conservation}).\\
(3) A statistical model is called {\it regular} if the set of 
true parameters
$
W_{0}=\{w\in W;q(x)=p(x|w)\}
$
consists of a single point and if the Fisher information matrix
is always positive definite. Note that a regular model 
is a very special example of singular learning machines. 
For a regular statistical model, we have
\begin{eqnarray*}
E[B^{*}_{g}]=\frac{d}{2},&&
E[G^{*}_{g}]=(1+\frac{1}{\beta})\frac{d}{2}, \\
E[B^{*}_{t}]=-\frac{d}{2},&&
E[G^{*}_{t}]=(-1+\frac{1}{\beta})\frac{d}{2},
\end{eqnarray*}
which is a special case of Theorem \ref{Theorem:222}.
\vskip3mm
Theorem \ref{Theorem:222} reveals
the universal relations among the four errors. It holds 
even if the set of true parameters has complex singularities. 
However, its statement simultaneously shows that we can extract no information 
about singularities directly from Theorem \ref{Theorem:222}.
Theorem \ref{Theorem:333}
shows that the four errors contain important information
about singularities. 
The Kullback-Leibler distance is
\[
K(w)=E_{X}[f(X,w)]=\int q(x)\log\frac{q(x)}{p(x|w)}dx.
\]
The {\it zeta function} of a learning machine
is defined by
\begin{equation}
\zeta (z)=\int_{W} K(w)^{z}\;\varphi(w)\;dw.
\label{eq:zeta}
\end{equation}
The zeta function is a holomorphic function of
a complex variable $z$ in the region
$Re(z)>0$, which can be analytically continued to
a meromorphic function on the entire complex plane. 
Its poles are all real, negative, and rational numbers
(for the proof, see \cite{Atiyah}\cite{Kashiwara}\cite{2001c}).
They are denoted as follows,
\[
0>-\lambda_{1}>-\lambda_{2}>-\lambda_{3}>\cdots .
\]
The order of each pole $\lambda_{k}$ is denoted by
$m_{k}$. We simply use notations
$\lambda=\lambda_{1}$ and $m=m_{1}$ for the largest pole
and its order respectively. 

\begin{Theorem}\label{Theorem:333}
As $n\rightarrow\infty$, the convergence in probability 
\[
nG_{g}+nG_{t}-\frac{2\lambda}{\beta}\rightarrow 0
\]
holds. Therefore
\begin{equation}
E[G^{*}_{g}]+E[G^{*}_{t}]=\frac{2\lambda}{\beta}.
\label{eq:gibbs-g-t}
\end{equation}
\end{Theorem}
Also the following corollary holds. 
\begin{Corollary}\label{Corollary:111}
The following convergence in probability holds,
\[
nB_{g}-nB_{t}+2nG_{t}-\frac{2\lambda}{\beta}\rightarrow 0.
\]
In particular, if $\beta=1$, $E[B^{*}_{g}]=\lambda$. 
\end{Corollary}
From these theorems and corollaries, if one knows the true
distribution, one can predict the Bayes and Gibbs generalization 
errors from the Bayes and Gibbs training errors with probability one,
as $n$ tends to infinity. In practical applications, we
seldom know the true distribution, however,
this fact is useful in computer simulation research of
learning theory and statistics. Lastly,  by
Theorems \ref{Theorem:222} and \ref{Theorem:333},
the following corollary is immediately proved. 
\begin{Corollary}
Let $\nu=\nu(\beta)=\beta (E[G_{t}^{*}]-E[B_{t}^{*}])$. Then
\begin{eqnarray*}
E[B_{g}^{*}]&=&\frac{\lambda-\nu}{\beta}+\nu,\\
E[B_{t}^{*}]&=&\frac{\lambda-\nu}{\beta}-\nu,\\
E[G_{g}^{*}]&=&\frac{\lambda}{\beta}+\nu,\\
E[G_{t}^{*}]&=&\frac{\lambda}{\beta}-\nu.
\end{eqnarray*}
Therefore Bayes learning is asymptotically determined by $\lambda$ and $\nu$. 
\end{Corollary}
In general $\nu(\beta)$ depend on $\beta>0$. 
In regular statistical models, $\lambda=\nu=d/2$ for arbitrary $\beta>0$, whereas
in singular learning machines, they are different in general. 
Corollary \ref{Corollary:111} was firstly 
discovered in \cite{1999}\cite{2001a}. 
Since the constant $\lambda$ depends strongly on the
true distribution, the learning machine, and 
the {\it a priori} distribution, it characterizes the
properties of 
learning machines. The values of several models 
have been studied in neural networks \cite{2001b},
normal mixtures \cite{Yamazaki1}, reduced rank regressions
\cite{Aoyagi}, Boltzmann machines \cite{Yamazaki2}, and
hidden Markov models \cite{Yamazaki3}. Also the 
behavior of $\lambda$ was analyzed for 
the case when Jeffreys' prior is employed as 
an {\it a priori} distribution \cite{2000}, and in the case when
the distance of the true distribution from the
singularity is in proportion to $1/\sqrt{n}$ \cite{2003}.

\section{Widely Applicable Information Criteria}

The main purpose of this paper is to prove 
 the theorems above. However, in order to illustrate 
the importance of the results of this paper,  we propose 
widely applicable information criteria and introduce an
experiment. Experimental analysis of practical
applications is a topic for future study. 

\subsection{Basic Concepts}

Based on Corollary \ref{Corollary:222}, we establish 
new information criteria which can be 
used for both regular and singular learning machines. 
Let us define the Bayes generalization loss, 
the Bayes training loss, 
the Gibbs generalization loss,
and the Gibbs training loss by 
\begin{eqnarray*}
BL_{g} & = & -E_{X}[\log E_{w}[p(X|w)] ],\\
BL_{t} & = & -\frac{1}{n}\sum_{j=1}^{n} \log E_{w}[p(X_{j}|w)] ,\\
GL_{g} &=& -E_{w}E_{X}[\log p(X|w)],\\
GL_{t} &=& -E_{w}[\frac{1}{n}\sum_{j=1}^{n}
\log p(X_{j}|w)].
\end{eqnarray*}
These losses are random variables. 
Both training losses $BL_{t}$ and $GL_{t}$ can 
be numerically calculated based on training samples $D_{n}$ 
and a learning machine $p(x|w)$ without any knowledge of the
true density function $q(x)$. By combining the entropy of the true 
distribution with Corollary \ref{Corollary:222},
\[
S=-\int q(x)\log q(x) dx = -
E\Bigl[\frac{1}{n}\sum_{i=1}^{n}\log q(X_{i})\Bigr],
\]
we obtain the equations, 
\begin{eqnarray*}
E[BL_{g}] &=& E[BL_{t}]+2\beta ( E[GL_{t}]-E[BL_{t}])+o(\frac{1}{n}), \\
E[GL_{g}] &=& E[GL_{t}]+2\beta ( E[GL_{t}]-E[BL_{t}])+o(\frac{1}{n}).
\end{eqnarray*}
Let us define widely applicable information criteria (WAIC) by
\begin{eqnarray*}
\mbox{WAIC}_{1} & = & BL_{t} + 2\beta\; (GL_{t}-BL_{t}), \\
\mbox{WAIC}_{2} & = & GL_{t} + 2\beta\; (GL_{t}-BL_{t}).
\end{eqnarray*}
Then the expectations of the two criteria respectively equal 
the Bayes and Gibbs generalization losses,
\begin{eqnarray*}
E[BL_{g}] &=& E[\mbox{WAIC}_{1}]+o(\frac{1}{n}),\\
E[GL_{g}] &=& E[\mbox{WAIC}_{2}]+o(\frac{1}{n}).
\end{eqnarray*}
Therefore, $\mbox{WAIC}_{1}$ and $\mbox{WAIC}_{2}$ provide 
indices for model evaluation. 
\vskip3mm\noindent
{\bf Remark.} If a model is regular and the true distribution is 
contained in the parametric model, then $\lambda=d/2$ and 
\begin{equation}
2\beta(E[G_{t}^{*}]-E[B_{t}^{*}])=d
\label{eq:AIC}
\end{equation}
hold. It is proved in \cite{NC2009-3} that,
even if a model $p(x|w)$ does not contain 
the true distribution $q(x)$, the equations of states hold
if the Hessian matrix of the
Kullback-Leibler distance is positive 
definite at the unique optimal paramater $w^{*}$
that minimizes the Kullback-Leibler distance
from $q(x)$ to $p(x|w)$. In such a case, 
\begin{equation}
2\beta(E[G_{t}^{*}]-E[B_{t}^{*}])=\mbox{tr}(IJ^{-1}),
\label{eq:TIC}
\end{equation}
where $I$ and $J$ are $d\times d$ matrices defined by
\begin{eqnarray*}
I_{ij}&=&\int\partial_{i}f(x,w^{*})\partial_{j} f(x,w^{*}) q(x)dx,\\
J_{ij}&=&-\int \partial_{i}\partial_{j} f(x,w^{*})q(x)dx.
\end{eqnarray*}
Here we used a notation, $\partial_{i}=(\partial/\partial w_{i})$. 
Moreover, as $n\rightarrow\infty$ convergence in probability
\begin{equation}
2\beta(G_{t}^{*}-B_{t}^{*})\rightarrow \mbox{tr}(IJ^{-1})
\label{eq:TIC-2}
\end{equation}
holds. If $\beta\rightarrow\infty$,
both the Bayes and Gibbs estimations result in 
the maximum likelihood method. Therefore, for regular statistical models,
WAIC has asymptotically the same variance as AIC. 
In other words, WAIC can be understood as 
information criteria of generalized from AIC. 
For singular learning machines, neither eq.(\ref{eq:AIC}) nor (\ref{eq:TIC})
holds, for example, $J^{-1}$ does not exist, whereas WAIC gives the 
accurate generalization error. 
\vskip3mm\noindent
{\bf Remark.} In Bayes estimation, the marginal likelihood
or the stochastic complexity
\[
F=-\log \int \varphi(w)\prod_{i=1}^{n}p(X_{i}|w)dw
\]
is often used in model selection and hyperparameter 
optimization. We clarified its behavior for singular
learning machines in \cite{2001a}. In regular statistical
models, $F$ is asymptotically equal to BIC, however, 
in singular models, it is not equal to BIC even asymptotically. 
Note that $F$ does not correspond to the generalization error,
hence the optimal model for the minimizing $F$ does not minimize 
the generalization error in general. The Bayes and Gibbs generalization errors
are important because they corresond directly to the Kullback-Leibler 
distance from the true distribution to the estimated one. 
In this paper, we make mathematically new information criteria
which correspond to the generalization error. 
Even for regular
statsitcal models, there is much research and discussion which compares
AIC with BIC. It is a topic for future study to compare the marginal likelihood 
and the equations of states from the viewpoint of statistical methodology. 
\vskip3mm\noindent
{\bf Remark.} In conventinal Bayes estimation, the inverse
temperature $\beta=1$ is used. Hence WAIC for 
$\beta=1$ is most important. On the other hand, WAIC for 
general $\beta$ shows the effect of the inverse temperature on
the generalization and training errors. 
Moreover, in applications, one may use $\beta$ as a hyperparameter.
In such a case, it can be optimized by the minimization of WAIC. 

\subsection{Experiments}

\begin{table}[tb]\label{table:111}
\begin{center}
\begin{tabular}{|c|c|c|c|c|c|}
\hline
$H$ & Theory & $E[B_{g}]$ & $\sigma[B_{g}]$ & $E[\mbox{WAIC}_{1}]$ & $\sigma[\mbox{WAIC}_{1}]$ \\
\hline
1 &          & 6.215318 & 0.034043 & 6.214185 & 0.230465 \\
2 &          & 3.013187 & 0.118109 & 2.993593 & 0.225722 \\
3 & 0.027000 & 0.028422 & 0.007393 & 0.025139 & 0.006886 \\
4 & 0.030000 & 0.030830 & 0.007678 & 0.027207 & 0.008176 \\
5 & 0.032000 & 0.033030 & 0.008418 & 0.030152 & 0.008728 \\
6 & 0.034000 & 0.034978 & 0.008832 & 0.031382 & 0.009778 \\
\hline
\end{tabular}
\end{center}
\caption{Experimental Results}
\end{table}

We studied reduced rank regressions. The input and output vector
is $x=(x_{1},x_{2})\in{\bf R}^{N_{1}}\times {\bf R}^{N_{2}}$
and the parameter is
$w=(A,B)$ where $A$ and $B$ are respectively $N_{1}\times H$ and 
$H\times N_{2}$ matrices. The learning machine is
\[
p(x|w)=q(x_{1})\frac{1}{(2\pi\sigma^{2})^{N_{2}/2}}
\exp(-\frac{1}{2\sigma^{2}}\|x_{2}-BAx_{1}\|^{2}).
\]
Since $q(x_{1})$ has no parameter, it is not estimated. 
The true distribution is determined by matrices $A_{0}$ and $B_{0}$ 
such that $\mbox{rank}(B_{0}A_{0})=H_{0}$. 
The algebraic variety of the true parameters is defined by $K(A,B)=0$, where
\[
K(A,B)\propto\|BA-B_{0}A_{0}\|^{2},
\]
has complicated singularities. 
We conducted experiments for the case that 
$N_{1}=N_{2}=6$, $H_{0}=3$, $\beta=1$, 
$n=500$, and $\sigma=0.1$. 
The {\it a priori} distribution was
$p(A,B)\propto \exp(-2.0\cdot10^{-5}(\|A\|^{2}+\|B\|^{2}))$. 
Reduced rank regressions with hidden units $H=1,2,..,6$ were 
employed. 
The {\it a posteriori} distribution was numerically approximated by 
the Metropolis method, where initial 5000 steps were omitted
and 2000 parameters were collected after every 200 steps. 
The expectation values $B_{g}$ and $WAIC_{1}$ were
obtained by averaging over 25 trials, that is to say, 25 sets of training samples were
independently taken from the true distribution. 
In Table.1, theoretical values of $E[B_{g}]$ 
for $\beta=1$ were obtained from \cite{Aoyagi}. Learning machines with $H=1,2$ do
not contain the true distribution, hence theoretical values do not exist.  
The two values 
$E[B_{g}]$ and $\sigma[B_{g}]$ are the experimental average and 
standard deviation of the Bayes generalization error, respectively. 
The two values
$E[\mbox{WAIC}_{1}]$ and $\sigma[\mbox{WAIC}_{1}]$ are the experimental average and 
standard deviation of $\mbox{WAIC}_{1}$, respectively. 
The experimental results show that the average behavior of 
the Bayes generalization error could be estimated by that of 
$\mbox{WAIC}_{1}$. However, 
the standard deviations of the $\mbox{WAIC}_{1}$ and 
the Bayes generalization error are not small. 
Note that, even in regular statistical models, the standard deviations of
the generalization error and AIC are also not small. 


\section{Singular Learning Theory}

\label{Section:proof}

In this section, we shall prove the main theorems. 
Proofs of the lemmas are rather technical, hence 
they are given in Appendix. 

\subsection{Outline of the Proof}

We prove the main theorems by the following procedure.\\
(1) Firstly we show that only 
the neighborhoods of the true parameters essentially 
affect the four errors. \\
(2) By using resolution of singularities,
the set of parameters can be understood as the image of an 
analytic map from a manifold, on which 
all singularities of the true parameters 
are of normal crossing type. \\
(3) We prove that the four errors converges in law to functionals of
a tight gaussian process on the set of true parameters
in the manifold. \\
(4) Expectations of the four errors converge to those
of functionals of the tight gaussian process. \\
(5) The relations between the four errors are 
derived by  partial integration of the gaussian process. 

\subsection{Basic Properties}

By using the log density ratio function 
$f(x,w)$, we define
the empirical Kullback-Leibler distance by
\[
K_{n}(w)=\frac{1}{n}\sum_{i=1}^{n}f(X_{i},w).
\]
For a given constant $a>0$, we define an expectation value 
restricted to the set $\{w\in W;K(w)\leq a\}$ by
\[
E_{w}[f(w)|_{K(w)\leq a}]=
\frac{\displaystyle
\int_{K(w)\leq a} f(w) e^{-\beta nK_{n}(w)} \varphi(w)dw}
{\displaystyle
\int_{K(w)\leq a} e^{-\beta nK_{n}(w)} \varphi(w)dw}.
\]
We define four errors respectively by
\begin{eqnarray*}
B_{g}(a)&=&
E_{X}\Bigl{[}-\log E_{w}[e^{-f(X,w)}|_{K(w)\leq a}]
\Bigr{]},\\
B_{t}(a)&=&
\frac{1}{n}\sum_{j=1}^{n}
-\log  E_{w}[e^{-f(X_{j},w)}|_{K(w)\leq a}],
\\
G_{g}(a)&=& E_{w}[K(w)|_{K(w)\leq a}],
\\
G_{t}(a)&=& E_{w}[K_{n}(w)|_{K(w)\leq a}].
\end{eqnarray*}
Since $W$ is compact and $K(w)$ is an analytic function,
$\overline{K}=\sup_{w\in W}K(w)$ is finite. 
Then,  $B_{g}(\overline{K})=B_{g}$, $B_{t}(\overline{K})=B_{t}$, 
$G_{g}(\overline{K})=G_{g}$, and $G_{t}(\overline{K})=G_{t}$.
Also we define $\eta_{n}(w)$ for $w$ such that $K(w)>0$ by
\begin{equation}
\eta_{n}(w)=\frac{K(w)-K_{n}(w)}
{\sqrt{K(w)}},\label{eq:eta}
\end{equation}
and
\[
H_{t}(a)=\sup_{0<K(w)\leq a}|\eta_{n}(w)|^{2}.
\]
$H_{t}(\overline{K})$ is denoted by $H_{t}$. 
\begin{Lemma} \label{Lemma:1}
For an arbitrary $a>0$, 
the following inequalities hold. 
\begin{eqnarray*}
&&B_{t}(a)\leq G_{t}(a)\leq
 \frac{3}{2}G_{g}(a)+ \frac{1}{2}H_{t}(a),\\
&&0\leq  B_{g}(a)\leq G_{g}(a),\\
&&-\frac{1}{4}H_{t}(a)\leq G_{t}(a).
\end{eqnarray*}
\end{Lemma}
For the proof of this lemma, see Section \ref{Chapter:Lemma}.
In particular, by putting $a=\overline{K}$, we have
\begin{eqnarray*}
&&B_{t}\leq G_{t}\leq \frac{3}{2}G_{g}+ \frac{1}{2}H_{t},\\
&&0\leq B_{g}\leq G_{g},\\
&&-\frac{1}{4}H_{t}\leq G_{t}.
\end{eqnarray*}

\noindent{\bf Remark.} A sequence of random variables $\{R_{n}\}$ is called 
asymptotically uniformly integrable (AUI) if
\[
\lim_{M\rightarrow\infty}\mbox{limsup}_{n\rightarrow\infty}
E[I_{M}(R_{n})]=0,
\]
where
\[
I_{M}(x)=\left\{
\begin{array}{cc}
0 &(|x|< M)\\
|x| &(|x|\geq M)
\end{array}\right..
\]
The following properties are well known \cite{Wvan}.\\ 
(1) If
the convergence in law $R_{n}\rightarrow R$ holds and 
$R_{n}$ is AUI, then $E[R_{n}]\rightarrow E[R]$. \\
(2) If $R_{n}$ is AUI and if a random variable $S_{n}$ satisfies 
$|S_{n}|\leq R_{n}$, then $S_{n}$ is also AUI. \\
(3) If there exist $p>0$ and $C>0$ such that $E[|R_{n}|^{p}]<C$, 
then $R_{n}^{q}$ $(0<q<p)$ is AUI.
\vskip3mm\noindent
By Lemma \ref{Lemma:1}, if 
$nH_{t}(a)$, $n G_{g}(a)$, and $nB_{t}(a)$ are AUI, then $nB_{g}(a)$ and 
$nG_{t}(a)$ are AUI. 
\begin{Lemma}\label{Lemma:2}
(1) There exists a constant $C_{H}>0$ such that
\[
E[(nH_{t})^{3}]=C_{H}<\infty.
\]
(2) For an arbitrary $\alpha>0$, 
\begin{equation}
\mbox{Pr}(nH_{t}>n^{\alpha})\leq \frac{C_{H}}{n^{3\alpha}}.
\label{eq:prob}
\end{equation}
\end{Lemma}
For the proof of this lemma, see Section \ref{Chapter:Lemma}. 
Lemma \ref{Lemma:2} shows that
$nH_{t}$ is asymptotically uniformly integrable.

\begin{Lemma}\label{Lemma:3}
(1) The four errors $n B_{g}$, $n B_{t}$, $n G_{g}$, and $n G_{t}$ are
all asymptotically uniformly integrable.\\
(2) For an arbitrary $\epsilon>0$, following convergences in probability hold
\begin{eqnarray*}
n(B_{g}-B_{g}(\epsilon))&\rightarrow & 0,\\
n(B_{t}-B_{t}(\epsilon))&\rightarrow & 0,\\
n(G_{g}-G_{g}(\epsilon))&\rightarrow & 0,\\
n(G_{t}-G_{t}(\epsilon))&\rightarrow & 0.
\end{eqnarray*}
\end{Lemma}
For the proof of this lemma, see Section \ref{Chapter:Lemma}.
Based on this Lemma, $B_{g}(\epsilon)$, $B_{t}(\epsilon)$,
$G_{g}(\epsilon)$, and $G_{t}(\epsilon)$ are referred to as
the major parts of the four errors. 

\subsection{Resolution of Singularities}

By Lemma\ref{Lemma:3}, the main region in the parameter set to be studied is 
\[
W_{\epsilon}=\{w\in W\;;\;K(w)\leq \epsilon\}
\]
for a sufficiently small $\epsilon >0$. 
By applying 
Hironaka's resolution theorem 
to $K(w)(\epsilon -K(w))\varphi_{1}(w)\pi_{1}(w)\cdots\pi_{k}(w)$, 
there exist a manifold ${\cal M}=\cup_{\alpha}U_{\alpha}$ 
where $U_{\alpha}$ is a local coordinate and
a proper analytic map $g:U_{\alpha}\rightarrow W_{\epsilon}$, 
expressed as $w=g(u)$, such that
in each $U_{\alpha}$, the functions
$K(w)$, $(\epsilon -K(w))$, $\varphi_{1}(w)$, $\pi_{1}(w)$, $ \cdots$, and $\pi_{k}(w)$
are all normal crossing. That is to say, 
\[
K(g(u))=u^{2k}=\prod_{j=1}^{d}u_{j}^{2k_{j}},
\]
and 
\[
\varphi(g(u))|g'(u)|= b(u) | u^{h}| 
= b(u) | \prod_{j=1}^{d}u_{j}^{h_{j}}|, 
\]
where $|g'(u)|$ is the Jacobian determinant, 
$k=(k_{1},k_{2},...,k_{d})$ and $h=(h_{1},h_{2},..,h_{d})$ 
are sets of nonnegative integers, and $b(u)>0$ is a $C^{\infty}$ class function. 
Note that $g(u)$, $k$, and $h$ depend on the local coordinate $U_{\alpha}$, however, 
to keep notation simple, we omit $\alpha$ that identifies the local coordinate. 
By applying partitions of unity to ${\cal M}$, we can assume that
$g^{-1}(W)$ is the union of coordinates $[0,1]^{d}$
and that
\[
\varphi(g(u))|g'(u)|=u^{h}\;\psi(u),
\]
where $\psi(u)>0$ is a $C^{\infty}$ class function, without loss of generality. 
Existence of such a manifold ${\cal M}$ and an analytic map $w=g(u)$ 
is well known in algebraic geometry \cite{AlgebraicG}, 
algebraic analysis\cite{Atiyah,Kashiwara}, and learning theory \cite{2001a}.  
Since $W_{\epsilon}$ is compact and $g$ is a proper map, $g^{-1}(W_{\epsilon})$ is
also compact. 
For our purpose, we need only the compact subset $g^{-1}(W_{\epsilon})$ 
in ${\cal M}$. Therefore, hereinafter we use the notation
${\cal M}$ for $g^{-1}(W_{\epsilon})$, which is a compact subset of the manifold. 
The set of true parameters is denoted by $W_{0}=\{w\in W\;;\;K(w)=0\}$ and
${\cal M}_{0}=\{u\in {\cal M}\;;\;K(g(u))=0\}$. 

Let us define the supremum norm by
\[
\|f\|=\sup_{u\in {\cal M}}|f(u)|.
\]
Then we have a standard form of the log density ratio function. 
\begin{Lemma}\label{Lemma:4}
There exists 
an $L^{s}(q)$ valued analytic function 
${\cal M}\ni u\mapsto a(x,u)\in L^{s}(q)$ such that 
\begin{eqnarray}
f(x,g(u))& = & a(x,u)\;u^{k}, \label{eq:Lemma4-1}\\
E_{X}[a(X,u)]& = & u^{k}, \label{eq:Lemma4-2}\\
K(g(u))=0&\Rightarrow& E_{X}[a(X,u)^{2}]=2, \label{eq:Lemma4-3}\\
E_{X}[\|a(X)\|^{s}]&<& \infty.\label{eq:Lemma4-4}
\end{eqnarray}
\end{Lemma}
This lemma shows that, if there are only normal crossing singularities
in the parameter set, the ideal generated by the set of true parameters 
is trivial, with the result that the log density ratio function is also trivial. 
For the proof of this lemma, see Section \ref{Chapter:Lemma}.
We define $\|a(X)\|=\sup_{u\in {\cal M}}|a(X,u)|$. 

\subsection{Empirical Processes}

An empirical process $\xi_{n}(u)$ 
is defined by
\[
\xi_{n}(u) = \frac{1}{\sqrt{n}}\sum_{i=1}^{n}
a^{*}(X_{i},u)
\]
where $a^{*}(x,u)=E_{X}[a(X,u)]-a(x,u)$. Note that
$|\xi_{n}(u)|=|\eta_{n}(g(u))|$, where $\eta_{n}(w)$ in eq.(\ref{eq:eta})
is ill-defined on $K(w)=0$ on $W$, but $\xi_{n}(u)$ is well-defined
on $K(g(u))=0$ on ${\cal M}$. In other words, resolution of singularities
ensures $\eta_{n}$ is well-defined. 
We have the following Lemma.
\begin{Lemma}\label{Lemma:5}
The empirical process satisfies
\begin{eqnarray*}  
E[ \|\xi_{n}\|^{6} ] & < & Const.<\infty \\
E[ \|\nabla\xi_{n}\|^{6}] &<& Const.<\infty
\end{eqnarray*}
where $Const.$ does not depend on $n$, and 
$\|\nabla \xi_{n}\|=\sum_{j=1}^{d}\|\partial_{j}\xi_{n}\|$. 
\end{Lemma}
Let the Banach space of uniformly bounded and continuous functions on
${\cal M}$ be
\[
B({\cal M})=\{f(u)\;;\;\|f\|<\infty\}.
\]
Since ${\cal M}$ is compact, $B({\cal M})$ is a separable normed space. 
It was proved in \cite{2005} that 
the empirical process $\xi_{n}(u)$ defined on $B({\cal M})$ weakly converges 
to the tight gaussian process $\xi(u)$ that satisfies
\begin{eqnarray*}
E_{\xi}[\xi(u)]& = & 0, \\
E_{\xi}[\xi(u)\xi(v)] &=& 
E_{X}[a^{*}(X,u)a^{*}(X,v)].
\end{eqnarray*}
If $u,v\in {\cal M}_{0}$, 
\[
E_{X}[a^{*}(X,u)a^{*}(X,v)]=E_{X}[a(X,u)a(X,v)].
\]
It is well known that a tight gaussian process is uniquely 
determined by its expectation and the covariance matrix of finite points. 
In a singular learning machine, the Fisher information matrix is singular,
however, $E_{X}[a(X,u)a(X,v)]$ can be understood as a generalized
version of the Fisher information matrix. 

Let $\xi(u)$ be an arbitrary differentiable function.
We define the average of $f(u)$ over 
${\cal M}$ for the given function $\xi(u)$ by
\[
E_{u}^{\sigma}[f(u)|\xi] = \frac{\displaystyle
\sum_{\alpha}\int_{[0,1]^{d}} f(u)\;Z(u,\xi)\;du
}{\displaystyle
\sum_{\alpha}\int_{[0,1]^{d}} Z(u,\xi)\;du
},
\]
where $\sum_{\alpha}$ is the sum over all coordinates
of ${\cal M}$, $\sigma$ is a constant which satisfies 
$0\leq \sigma \leq 1$, and
\[
Z(u,\xi)= u^{h}\;\psi(u)\;
e^{-\beta nu^{2k}+\beta \sqrt{n}u^{k}\xi(u)+\sigma u^{k} a(X,u)}.
\]
\begin{Lemma}\label{Lemma:6} Assume that $k_{1}>0$. 
For an arbitrary analytic function $\xi(u)$,
\begin{eqnarray*}
E_{u}^{\sigma}[u^{2k}|\xi]
& \leq & \frac{c_{1}}{n}\{1+
\|\xi\|^{2}
+\|\partial_{1}\xi\|^{2}\\
&&+\sigma\|a(X)\|+\sigma\|\partial_{1} a(X)\|\}, \\
E_{u}^{\sigma}[u^{3k}|\xi]
& \leq & \frac{c_{2}}{n^{3/2}}\{1+
\|\xi\|^{3}
+\|\partial_{1}\xi\|^{3}\\
&& +(\sigma\|a(X)\|)^{3/2}+(\sigma\|\partial_{1} a(X)\|)^{3/2}\},
\end{eqnarray*}
where $\partial_{1}=(\partial/\partial u_{1})$, and $c_{1},c_{2},c_{3}>0$ are
constants which are determined by $k_{1},h_{1}$, $\beta$, and $\|\psi\|\|1/\psi\|$. 
\end{Lemma}
Note that, by Lemma \ref{Lemma:6}, $G_{g}(\epsilon)$ is asymptotically 
uniformly integrable. 
For the proof of this Lemma, see Section \ref{Chapter:Lemma}.

Since $w=g(u)$, we rewrite the major parts of four errors by using the emprical 
process $\xi_{n}(u)$, 
\begin{eqnarray}
B_{g}(\epsilon)&= &E_{X}[-\log E_{u}^{0}[e^{-a(X,u)u^{k}}|\xi_{n}]] ,\label{eq:bg-1}\\
B_{t}(\epsilon)&=&\frac{1}{n}\sum_{j=1}^{n}-\log E_{u}^{0}
[e^{-a(X_{j},u)u^{k}}|\xi_{n}],\label{eq:bt-1}\\
G_{g}(\epsilon)&=&E_{u}^{0}[ u^{2k}|\xi_{n}], \label{eq:gg-1}\\
G_{t}(\epsilon)&=&E_{u}^{0}[
u^{2k}-\frac{1}{\sqrt{n}}\;u^{k}\xi_{n}(u)|\xi_{n}].\label{eq:gt-1}
\end{eqnarray}
In each local coordinate $[0,1]^{d}$, without loss of generality, we can assume that
there exists $r$ such that 
\[
u=(x,y)\in {\bf R}^{r}\times {\bf R}^{r'}, 
\]
where $ r'=d-r$, multi-indeces $k=(k,k')$ and $h=(h,h')$ satisfy
\[
\frac{h_{1}+1}{2k_{1}}
= \cdots=\frac{h_{r}+1}{2k_{r}}=\lambda_{\alpha}
<\frac{h'_{1}+1}{2k'_{1}}\leq \cdots,
\]
where $(-\lambda_{\alpha})$ and $r$ are respectively equal to the largest pole 
and its order of 
the meromorphic function that is given by the analytic continuation of 
\[
\int_{[0,1]^{d}}u^{2kz+h}du.
\]
We define the multi-index $\mu=(\mu_{1},...,\mu_{r'})\in {\bf R}^{r'}$ by
\[
\mu_{i}=h'_{i}-2k'_{i}\lambda_{\alpha}.
\]
Then
\[
\mu_{i}> h'_{i}-2k'_{i}\Bigl(\frac{h'_{i}+1}{2k'_{i}}\Bigr)=-1,
\]
hence $y^{\mu}$ is integrable in $[0,1]^{r'}$. Both $\lambda_{\alpha}$ and $r$ 
depend on the local coordinate. Let $\lambda$ be the 
smallest $\lambda_{\alpha}$, and $m$ be the largest 
$r$ among the coordinates for which $\lambda=\lambda_{\alpha}$. 
Then $(-\lambda)$ and $m$ are respectively equal to 
the largest pole and its order of the zeta function 
of eq.(\ref{eq:zeta}).
Let $\alpha^{*}$ be the index of the set of all
coordinates that satisfy
$\lambda_{\alpha}=\lambda$ and $r=m$. As is shown by
the following lemma, only the coordinates $U_{\alpha^{*}}$ 
affect the four errors. Let $\sum_{\alpha^{*}}$ denote
the sum over all such coordinates. 

For a given function $f(u)$, we adopt the notation $f_{0}(y)=f(0,y)$.
For example, $a_{0}(X,y)=a(X,0,y)$, $\xi_{0}(y)=\xi(0,y)$, and 
$\psi_{0}(y)=\psi(0,y)$. 
The expectation value for a given function
$\xi(u)$ is defined by
\[
E_{y,t}[f(y,t)|\xi]=
\frac{\displaystyle
\sum_{\alpha^{*}}\int_{0}^{\infty} dt \int dy\;
f(y,t)\;Z_{0}(y,t,\xi)}{
\displaystyle
\sum_{\alpha^{*}}
\int_{0}^{\infty}dt \int dy\;
Z_{0}(y,t,\xi)}
\]
where $\int dy$ denotes $\int_{[0,1]^{r'}}dy$ and 
\[
Z_{0}(y,t,\xi)=y^{\mu}\;t^{\lambda-1}
e^{-\beta t+\beta \sqrt{t}\;\xi_{0}(y)}\psi_{0}(y).
\]
Then we have the following lemma. 
\begin{Lemma} \label{Lemma:7}
Let $p\geq 0$ be a constant. 
There exists $c_{1}>0$ such that,
for an arbitrary $C^{1}$-class function $f(u)$ 
and analytic function $\xi(u)$, 
the following inequality holds,
\begin{eqnarray*}
\left.
\begin{array}{l}
\displaystyle
\Big{|}\;n^{p}\;E_{u}^{0}[u^{2pk}f(u)|\xi]
-\;E_{y,t}[t^{p}f_{0}(y)|\xi]
\Big{|} \\
\leq \displaystyle \frac{c_{1}}{\log n}
\exp(4\beta\|\xi\|^{2})
 \{ \beta \|\nabla \xi\|\|f\|+\|\nabla f\|+\|f\|\}
\end{array}\right.
\end{eqnarray*}
where $\|\nabla f\|=\sum_{j}\|\partial_{j}f\|$.
\end{Lemma}
We define four functionals of a  given function $\xi(u)$ by
\begin{eqnarray}
B_{g}^{*}(\xi)&\equiv &\frac{1}{2}
E_{X}[\;E_{y,t}[a_{0}(X,y)t^{1/2}|\xi]^{2}\;] ,\label{eq:bg-2}\\
B_{t}^{*}(\xi)&\equiv & G^{*}_{t}(\xi)-G^{*}_{g}(\xi)+B^{*}_{g}(\xi),\label{eq:bt-2}\\
G_{g}^{*}(\xi)&\equiv &E_{y,t}[ t|\xi], \label{eq:gg-2}\\
G_{t}^{*}(\xi)&\equiv &E_{y,t}[t-t^{1/2}\xi_{0}(y)|\xi].\label{eq:gt-2}
\end{eqnarray}
Note that these four functionals do not depend on $n$. 
From the definition, we can prove the following lemma. 

\begin{Lemma}\label{Lemma:8}
For an arbitrary real measurable function $\xi(u)$,
\[
G_{g}^{*}(\xi)+G_{t}^{*}(\xi)=\frac{2\lambda}{\beta}.
\]
\end{Lemma}

\subsection{Proof of Theorem \ref{Theorem:111}}

Firstly we show that the following convergences in probability hold.
\begin{eqnarray}
nB_{g}(\epsilon)-B_{g}^{*}(\xi_{n})& \rightarrow& 0,\label{eq:pr1}\\
nB_{t}(\epsilon)-B_{t}^{*}(\xi_{n})& \rightarrow& 0,\label{eq:pr2}\\
nG_{g}(\epsilon)-G_{g}^{*}(\xi_{n})& \rightarrow& 0,\label{eq:pr3}\\
nG_{t}(\epsilon)-G_{t}^{*}(\xi_{n})& \rightarrow& 0.\label{eq:pr4}
\end{eqnarray}
Based on eq.(\ref{eq:gg-1}) and eq.(\ref{eq:gg-2}), we obtain
eq.(\ref{eq:pr3}) by Lemma \ref{Lemma:7}.
Also based on eq.(\ref{eq:gt-1}) and eq.(\ref{eq:gt-2}), we obtain
eq.(\ref{eq:pr4}) by Lemma \ref{Lemma:7}.
To prove eq.(\ref{eq:pr1}), we define 
\[
b_{g}(\sigma)\equiv E_{X}\Bigl[-\log E_{u}^{0}
[e^{-\sigma a(X,u)u^{k}}|\xi_{n}]\;\Bigr],
\]
then, it follows that 
 $nB_{g}(\epsilon)=nb_{g}(1)$ and there exists $0<\sigma^{*}<1$ such that
\begin{eqnarray}
nB_{g}(\epsilon)& = & n E_{u}^{0}[u^{2k}|\xi_{n}] 
- \frac{n}{2}E_{X}
E_{u}^{0}[a(X,u)^{2}u^{2k}|\xi_{n}] \nonumber \\
& & + \frac{n}{2}E_{X}
E_{u}^{0}[a(X,u)u^{k}|\xi_{n}]^{2}
+ \frac{1}{6}nb_{g}^{(3)}(\sigma^{*}),\label{eq:bgbg}
\end{eqnarray}
where we have used $E_{X}[a(X,u)]=u^{k}$. The 
first term on the right hand side of eq.(\ref{eq:bgbg}) 
is $nG_{g}(\epsilon)$. 
By Lemma \ref{Lemma:7}, we can prove 
the convergence in probability
\begin{eqnarray}
&&\Bigl|
n E_{X}E_{u}^{0}[a(X,u)^{2}u^{2k}|\xi_{n}] 
- E_{X}E_{y,t}[a_{0}(X,y)^{2}t|\xi_{n}]
\Bigr| \nonumber \\
&\leq & 
\frac{c_{1}}{\log n}e^{4\beta\|\xi_{n}\|^{2}}
E_{X}[\;\beta\|\nabla\xi_{n}\|\|a(X)^{2}\|+\|\nabla a(X)^{2}\|+\|a(X)^{2}\|\;]
\rightarrow 0\label{eq:bgt-0}
\end{eqnarray}
holds. The proof of eq.(\ref{eq:bgt-0}) is as follows. 
Two empirical processes $\xi_{n}(u)$ and $\partial \xi(u)$ 
respectively converge in law to $\xi(u)$ and $\partial \xi(u)$ 
in the Banach space with the sup norm $\|\;\;\|$. Therefore, their continuous functionals
$\|\xi_{n}\|$, $\|\partial \xi_{n}\|$, and $e^{4\beta\|\xi_{n}\|^{2}}$ 
also converge in law. Note that $1/\log n$ goes to zero. 
In general, if a sequence of random variables converges to zero in law,
then it converges to zero in probability, hence we obtain the convergence
in probability eq.(\ref{eq:bgt-0}).  In the following proofs, we use the 
same method. 

Since $E_{X}[a_{0}(X,y)]=2$, the sum of the 
first two terms of the right hand side of
eq.(\ref{eq:bgbg}) converges to zero in probability. For the third term, 
by using the notation
$E_{X}[a(X,u)a(X,v)]=\rho(u,v)$, $\rho_{0}(u,y)=\rho(u,(0,y))$,
and $\rho_{00}(y',y)=\rho((0,y'),(0,y))$, and applying Lemma \ref{Lemma:7},
\begin{eqnarray}
&&|nE_{X}E_{u}^{0}[a(X,u)u^{k}|\xi_{n}]^{2}
- E_{y,t}[a_{0}(X,y)t^{1/2}|\xi_{n}]^{2}| \nonumber \\
&\leq &
\Bigl| \sqrt{n}E_{u}^{0}\Bigl[u^{k}(\sqrt{n}E_{v}^{0}[\rho(u,v)v^{k}]
-E_{y,t}[\rho_{0}(u,y)t^{1/2}])\Bigr] \Bigr|\nonumber \\
&&+\Bigl|E_{y,t}\Bigl[t^{1/2}(\sqrt{n}E_{u}^{0}[\rho_{0}(u,y)u^{k}]
- E_{y',t'}[\rho_{00}(y',y)(t't)^{1/2}])\Bigr]\Bigr| \nonumber \\
&\leq & 
\frac{c_{1}\sqrt{n}}{\log n}
E_{u}^{0}[u^{k}]\;e^{4\beta\|\xi_{n}\|^{2}}(\beta\|\nabla\xi_{n}\|
\|\rho\|+\|\nabla\rho\|+\|\rho\|) \nonumber \\
&&+ \frac{c_{1}}{\log n} e^{4\beta\|\xi_{n}\|^{2}}
(\beta\|\nabla\xi_{n}\|\|\rho\|
+\|\nabla \rho\|+\|\rho\|),
\label{eq:bgt-1}
\end{eqnarray}
where `$|\xi_{n}$' is omitted to keep the notation simple. 
The equation (\ref{eq:bgt-1}) converges to zero in probability
by Lemma \ref{Lemma:6}.
Therefore the difference between the third term and $B_{g}^{*}(\xi_{n})$ 
converges to zero in probability. For the last term, we have
\begin{eqnarray*}
|nb^{(3)}(\sigma^{*})|
& = & \bigl|
E_{X}\Bigl\{
E_{u}^{\sigma^{*}}
[a(X,u)^{3}u^{3k}|\xi_{n}]
+2
E_{u}^{\sigma^{*}}
[a(X,u)|\xi_{n}]^{3}\\
& & 
-3 E_{u}^{\sigma^{*}}
[a(X,u)^{2}u^{2k}|\xi_{n}]
E_{u}^{\sigma^{*}}
[a(X,u)u|\xi_{n}]\Bigr\}\Bigr|
\\
&\leq & 
6 n  E_{X}\Bigl[\|a(X)\|^{3}\;E_{u}^{\sigma^{*}}
[u^{3k}|\xi_{n}]\Bigr].
\end{eqnarray*}
By applying Lemma \ref{Lemma:6},
\begin{eqnarray}
|nb_{g}^{(3)}(\sigma^{*})|
& \leq &
\frac{6 c_{2}}{n^{1/2}}E_{X}\Bigl[\|a(X)\|^{3}\;
\{1+\|\xi_{n}\|^{3}+\|\partial\xi_{n}\|^{3}
\nonumber \\
&&+\|a(X)\|^{3/2} +\|\partial a(X)\|^{3/2}
\}\Bigr], \label{eq:bgt-2}
\end{eqnarray}
which shows that $nb_{g}^{(3)}(\sigma^{*})$ converges to zero in probability.
Hence eq.(\ref{eq:pr1}) is proved. 
Let us prove eq.(\ref{eq:pr2}). By defining
\[
b_{t}(\sigma)=\frac{1}{n}\sum_{j=1}^{n}-\log E_{u}^{0}
[e^{-\sigma a(X_{j},u)u^{k}}|\xi_{n}],
\]
it follows that 
 $nB_{t}(\epsilon)=nb_{t}(1)$ and there exists $0<\sigma^{*}<1$ such that
\begin{eqnarray*}
nB_{t}(\epsilon)& = & n G_{t}(\epsilon)
- \frac{1}{2}\sum_{j=1}^{n}
E_{u}^{0}[a(X_{j},u)^{2}u^{2k}|\xi_{n}]  \\
& & + \frac{1}{2}\sum_{j=1}^{n}
E_{u}^{0}[a(X_{j},u)u^{k}|\xi_{n}]^{2}
+ \frac{1}{6}nb_{t}^{(3)}(\sigma^{*}),
\end{eqnarray*}
Then by applying Lemma \ref{Lemma:6}, $nb_{t}^{(3)}(\sigma^{*})$ 
converges to zero in probability in the same way as for eq.(\ref{eq:bgt-2}).
By the same methods as used with eq.(\ref{eq:bgt-0}) and eq.(\ref{eq:bgt-1}),
replacing respectively $E_{X}[\|a(X)^{2}\|]$ and $\rho(u,v)$ with
$(1/n)\sum_{j}\|a(X_{j})^{2}\|$ and 
$\rho_{n}=(1/n)\sum_{j}a(X_{j},u)a(X_{j},v)$,
convergences in probability
\begin{eqnarray*}
\frac{1}{2}\sum_{j=1}^{n}
E_{u}^{0}[a(X_{j},u)^{2}u^{2k}|\xi_{n}] 
- G_{g}^{*} (\xi_{n}) &\rightarrow 0 \\
\frac{1}{2}\sum_{j=1}^{n}
E_{u}^{0}[a(X_{j},u)u^{k}|\xi_{n}]^{2}
- B_{g}^{*}(\xi_{n})
&\rightarrow & 0
\end{eqnarray*}
hold, with the result that the convergence in probability
\begin{equation}
nB_{t}(\epsilon)-nG_{t}(\epsilon)+nG_{g}(\epsilon)-nB_{g}(\epsilon)\rightarrow 0.
\label{eq:four}
\end{equation}
holds. Therefore eq.(\ref{eq:pr2}) is obtained. 
By combining eq.(\ref{eq:pr1})-eq.(\ref{eq:pr4}) with Lemma \ref{Lemma:3} (2),
the following convergences in probability hold,
\begin{eqnarray}
nB_{g}-B_{g}^{*}(\xi_{n})& \rightarrow& 0,\\
nB_{t}-B_{t}^{*}(\xi_{n})& \rightarrow& 0,\\
nG_{g}-G_{g}^{*}(\xi_{n})& \rightarrow& 0,\\
nG_{t}-G_{t}^{*}(\xi_{n})& \rightarrow& 0.
\end{eqnarray}
Four functionals $B_{g}^{*}(\xi)$, $B_{t}^{*}(\xi)$, $G_{g}^{*}(\xi)$,
and $G_{t}^{*}(\xi)$ are continuous functions of 
$\xi\in B({\cal M})$. From the convergence in law of the empirical process 
$\xi_{n}\rightarrow\xi$, the convergences in law
\begin{eqnarray*}
B_{g}^{*}(\xi_{n})\rightarrow B_{g}^{*}(\xi),&&
B_{t}^{*}(\xi_{n}) \rightarrow B_{t}^{*}(\xi),\\
G_{g}^{*}(\xi_{n}) \rightarrow G_{g}^{*}(\xi),&&
G_{t}^{*}(\xi_{n})\rightarrow G_{t}^{*}(\xi)
\end{eqnarray*}
are derived. Therefore Theorem \ref{Theorem:111} (1) and (2) are obtained. 
Theorem \ref{Theorem:111} (3) is shown in Lemma \ref{Lemma:3}. (Q.E.D.)

\subsection{Proof of Theorem \ref{Theorem:222}}

Let $\{(x_{i},g_{i});i=1,2,...,N\}$ be a set of independent random variables
which are subject to the probability distribution
\[
q(x)\;\frac{e^{-g^{2}/2}}{\sqrt{2\pi}}. 
\]
A tight gaussian process is defined by
\[
\zeta_{n}(u)=\frac{1}{\sqrt{n}}\sum_{i=1}^{n}a(x_{i},u)g_{i}.
\]
Then, in the same way as the convergence in law $\xi_{n}(u)\rightarrow \xi(u)$ was proved, 
the covergence in law $\zeta_{n}(u)\rightarrow\xi(u)$ can be proved, 
because $\zeta_{n}(u)$ has the same expectation and covariance. 
\begin{eqnarray}
E[\zeta_{n}(u)]&=&0,\label{eq:xi-aa},\\
E[\zeta_{n}(u)\zeta_{n}(v)]&=&E_{X}[a(X,u)a(X,v)].\label{eq:xixi-aa}
\end{eqnarray}
In other words, both $\zeta_{n}(u)$ and $\xi_{n}(u)$ converge in law to
the same random process $\xi(u)$. Moreover, we can prove that $\zeta_{n}(u)$ satisfies
$E[\|\zeta_{n}\|^{s}]<\infty$ ($s\geq 6$) in the same way. 
Therefore we can prove equations of a gaussian random process $\xi(u)$ by
using the convergence in law $\zeta_{n}(u)\rightarrow \xi(u)$. 
Since $g_{i}$ is subject to the standard
normal distribution, 
\begin{equation}
E[g_{i}F(g_{i})]=E[\frac{\partial}{\partial g_{i}}F(g_{i})]
\label{eq:gF(g)}
\end{equation}
holds for 
a differentiable function of $F(x)$ which satisfies $|F(x)|/|x|^{k},|F(x)'|/|x|^{k}\rightarrow 0$ 
$(|x|\rightarrow\infty)$ for some $k>0$. 

Let us prove Theorem \ref{Theorem:222}. We use the notation,
\begin{eqnarray*}
Y(a) & = & \int_{0}^{\infty}dt\;
t^{\lambda-1}\;e^{-\beta t+a\beta\sqrt{t}},\\
\int du^{*} &=& \sum_{\alpha^{*}}\int dx\;dy\;\delta(x)\;y^{\mu}, \\
Z(\xi)&=&\int du^{*}\;Y(\xi(u)),
\end{eqnarray*}
where $u=(x,y)$. Also we define the expectation value 
of $f(u,t)$ for a given function $\xi(u)$, 
\[
\langle f(u,t) \rangle_{\xi}
=\frac{\int du^{*}
\int_{0}^{\infty}dt\; f(u,t)\;
t^{\lambda-1}\;e^{-\beta t+\xi(u)\beta\sqrt{t}}}
{
\int du^{*}\int_{0}^{\infty}dt\;
t^{\lambda-1}\;e^{-\beta t+\xi(u)\beta\sqrt{t}}}.
\]
Note that Lemma \ref{Lemma:8} is equivalent to 
\[
\langle 2t \rangle_{\xi} -\langle \sqrt{t}\xi(u)\rangle_{\xi}=\frac{2\lambda}{\beta}.
\]
By this equation and $|\sqrt{t}\xi(u)|\leq (t+\xi(u)^{2})/2$, 
\begin{eqnarray}
\langle t\rangle_{\xi} &\leq & 
\frac{4\lambda}{3\beta}+\frac{\langle\xi(u)^{2}\rangle_{\xi}}{3},\label{eq:ltr}\\
\langle |\sqrt{t}\xi(u)|\rangle_{\xi} &\leq &
\frac{2\lambda}{3\beta}+\frac{2\langle\xi(u)^{2}\rangle_{\xi}}{3},\label{eq:ltxr}
\end{eqnarray}
hold for an arbitrary function $\xi(u)$. 
Note that $\langle \xi(u)^{2}\rangle_{\xi}\leq \|\xi\|^{2}$,
because $\|\;\;\|$ is the sup norm. 
The expectations of $B^{*}_{g}$, $G^{*}_{g}$, and $G^{*}_{t}$ can be written by
\begin{eqnarray*}
2E[B^{*}_{g}]&=&\frac{1}{\beta^{2}}E[E_{X}[
\Bigl{(}
\frac{\int du^{*} a(X,u)Y'(\xi(u))}
{Z(\xi)}
\Bigr{)}^{2}]], \\
E[G^{*}_{g}]&=&\frac{1}{\beta^{2}}
E[\frac{\int du^{*} Y''(\xi(u))}
{Z(\xi)}],\\
E[G^{*}_{t}]&=&
\frac{1}{\beta^{2}}E[
\frac{\int du^{*} 
Y''(\xi(u))}{Z(\xi)}]
- \frac{A}{\beta},
\end{eqnarray*}
where $A$ is a constant defined by
\[
A \equiv 
E[
\frac
{\int du^{*}\;\xi(u)Y'(\xi(u))
}{Z(\xi)}].
\]
We introduce $A_{n}$ by using $\zeta_{n}(u)$,
\begin{eqnarray*}
A_{n}
&=&
E[\frac{
\int du^{*}\;\zeta_{n}(u)Y'(\zeta_{n}(u))}{Z(\zeta_{n})}]\\
&=&\beta E[\langle \zeta_{n}\sqrt{t}\rangle_{\zeta_{n}}]\\
&=&\frac{1}{\sqrt{n}}\sum_{i=1}^{n}\beta E[g_{i}\langle a(x_{i},u)\sqrt{t}\rangle_{\zeta_{n}}].
\end{eqnarray*}
Then by eq.(\ref{eq:ltxr}), $\langle \zeta_{n}(u)\sqrt{t}\rangle_{\zeta_{n}}$ is
asymptotically uniformly integrable, hence $A_{n}\rightarrow A$ $(n\rightarrow\infty)$. 
On the other hand, we define
\begin{eqnarray*}
B_{n}&=&\frac{1}{\sqrt{n}}
\sum_{i=1}^{n}\beta E[\frac{\partial}{\partial g_{i}}
\langle a(x_{i},u)\sqrt{t}g_{i}\rangle_{\zeta_{n}}]\\
&=& 
E[
\int du^{*}\{\frac{1}{\sqrt{n}}\sum_{i=1}^{n}
a(x_{i},u)\frac{\partial}{\partial g_{i}}\}\frac{Y'(\zeta_{n}(u))}{Z(\zeta_{n})}].
\end{eqnarray*}
Then by using
\begin{eqnarray*}
\frac{\partial}{\partial g_{i}}\Bigl(\frac{Y'(\zeta_{n}(u))}{Z(\zeta_{n})}\Bigr)
&=&\frac{Y''(\zeta_{n}(u))a(x_{i},u)}{\sqrt{n}\;Z(\zeta_{n})}\\
&&-\frac{Y'(\zeta_{n}(u))}{\sqrt{n}\;Z(\zeta_{n})^{2}}
\int dv^{*}\; Y'(\zeta_{n}(v))a(x_{i},v),
\end{eqnarray*}
we have
\[
\beta \frac{\partial}{\partial g_{i}}
\langle a(x_{i},u)\sqrt{t}\rangle_{\zeta_{n}}
=\frac{\beta^{2}}{\sqrt{n}}\Bigl(
\langle a(x_{i},u)^{2}t\rangle_{\zeta_{n}}
-\langle a(x_{i},u)\sqrt{t}\rangle_{\zeta_{n}}^{2}\Bigr).
\]
Hence
\[
B_{n}=E[
\frac{\beta^{2}}{n}\sum_{i=1}^{n}\Bigl(
\langle a(x_{i},u)^{2}t\rangle_{\zeta_{n}}
-\langle a(x_{i},u)\sqrt{t}\rangle_{\zeta_{n}}^{2}\Bigr)].
\]
Also,
\begin{equation}\label{eq:upp_bd}
\zeta_{n}(u)^{2}\leq \frac{1}{n}
(\sum_{k=1}^{n}a(x_{i},u)^{2})
(\sum_{k=1}^{n}g_{i}^{2}).
\end{equation}
From eq.(\ref{eq:ltr}), eq.(\ref{eq:ltxr}), and eq.(\ref{eq:upp_bd}),
both $\langle a(x_{i},u) \sqrt{t}\rangle_{\zeta_{n}}$ 
and $(\partial/\partial g_{i})
\langle a(x_{i},u)\sqrt{t}\rangle_{\zeta_{n}}$ are 
bounded by a finite sum of quadratic forms of $g_{i}$. 
Hence by eq.(\ref{eq:gF(g)}), $A_{n}=B_{n}$. Lastly, since
$\langle (1/n)\sum_{i=1}^{n}a(x_{i},u)^{2}t\rangle_{\zeta_{n}}$ and
$\langle (1/\sqrt{n})\sum_{i=1}^{n}a(x_{i},u)\sqrt{t}\rangle_{\zeta_{n}}^{2}$
are asymptotically uniformly 
integrable by eq.(\ref{eq:ltr}), eq.(\ref{eq:ltxr}),
we obtain $B_{n}\rightarrow B$, where
\begin{eqnarray*}
B &=& E[
\int du^{*}
\frac{2Y''(\xi(u))
}{Z(\xi)}] 
-E_{X}[
\Bigl{(}
\frac{\int du^{*} \; a(X,u)Y'(\xi(u))}
{Z(\xi)}
\Bigr{)}^{2}]\\
&=& 2\beta^{2}E[G^{*}_{g}]-2\beta^{2}E[B^{*}_{g}].
\end{eqnarray*}
Here we have used $E_{X}[a(X,u)^{2}]=2$ for $K(g(u))=0$ by Lemma \ref{Lemma:4}.
Since $A_{n}=B_{n}$, $A_{n}\rightarrow A$, and $B_{n}\rightarrow B$, we have $A=B$. Therefore
\[
A=\beta (E[G^{*}_{g}]-E[G_{t}^{*}]),
\]
which completes Theorem \ref{Theorem:222}.
(Q.E.D.)

\subsection{Proof of Theorem \ref{Theorem:333}}

From Lemma \ref{Lemma:8}, 
it follows that
\[
G_{g}^{*}(\xi_{n})+G_{t}^{*}(\xi_{n})=\frac{2\lambda}{\beta}.
\]
Then by Theorem \ref{Theorem:111} and Lemma \ref{Lemma:3},
we obtain Theorem
\ref{Theorem:333}.
(Q.E.D.)

\section{Discussion}

In this section, we discuss the theorems in this paper. 

Firstly, Theorem \ref{Theorem:111} was derived from 
definitions of the four errors. As is shown in the proof,
\[
B_{t}=G_{t}-\hat{G}_{g}+\hat{B}_{g}+o_{p}(\frac{1}{n}),
\]
where $o_{p}(1/n)$ is a random variable whose order is
smaller than $1/n$ and
\begin{eqnarray*}
\hat{G}_{g}&=&
\frac{1}{2n}\sum_{j=1}^{n}E_{w}\Bigl[
(\log\frac{q(X_{j})}{p(X_{j}|w)})^{2}\Bigr], \\
\hat{B}_{g}&=&
\frac{1}{2n}\sum_{j=1}^{n}E_{w}[
\log\frac{q(X_{j})}{p(X_{j}|w)} ]^{2}.
\end{eqnarray*}
Here convergences in probability $n(\hat{G}_{g}-G_{g})\rightarrow 0$
and $n(\hat{B}_{g}- B_{g})\rightarrow 0$ hold. We need the information about
the true distribution to calculate both $\hat{G}_{g}$ and 
$\hat{B}_{g}$, however, we do not need it to calculate
\[
V\equiv 2(\hat{G}_{g}-\hat{B}_{g})
= 
\frac{1}{n}\sum_{j=1}^{n}E_{w}[
(\log p(X_{j}|w))^{2}]
-
\frac{1}{n}\sum_{j=1}^{n}E_{w}[
\log p(X_{j}|w)]^{2}.
\]
The random variable $V$ is the variance of the {\it a posteriori}
distribution. By using $V$, $WAIC_{1}$ and $WAIC_{2}$ can be replaced by
\begin{eqnarray*}
WAIC_{1}&=&BL_{t}+\beta V,\\
WAIC_{2}&=&GL_{t}+\beta V.
\end{eqnarray*}
The third criterion $WAIC_{3}$ 
\[
WAIC_{3}
= BL_{t}-GL_{t}+\hat{G}_{g}-\hat{B}_{g}
\]
can be used as an index to examine how precisely the asymptotic theory
holds. In other words, the value $|WAIC_{3}|$ 
is the error of the asymptotic theory. 

Secondly, let us study Theorem \ref{Theorem:222}.
This theorem is essentially derived from
the fact that the empirical process $\xi_{n}(u)$ 
converges to the tight gaussian process $\xi(u)$ and that
the partial integration formula 
\[
E[g_{i}F(g)]=E[\frac{\partial}{\partial g_{i}}F(g)]
\]
holds for $\xi(u)$. 

Thirdly, Theorem \ref{Theorem:333} is proved by 
the property of the integral 
\[
Z_{\lambda}(\beta|a)=\sum_{\alpha^{*}}\int du^{*}
\int_{0}^{\infty}dt\;t^{\lambda-1}\;e^{-\beta t+a\beta\sqrt{t}}.
\]
That is to say, Theorems \ref{Theorem:222} and \ref{Theorem:333}
are essentially proved by partial integration.

Fourthly, in this paper, we proved three results 
eqs.(\ref{eq:ES1}), (\ref{eq:ES2}), and (\ref{eq:gibbs-g-t}). 
The two relations of eq.(\ref{eq:ES1}) and eq.(\ref{eq:ES2})  hold
universally, independently of singularities, whereas 
the third relation of eq.(\ref{eq:gibbs-g-t}) depends strongly on 
singularities. To determine the values of the four errors,
one more relation is needed. However, it seems that
there is no such relation. Hence in order to 
determine the four errors, we may have to evaluate
at least one of the four errors. For example
\[
E[G_{t}]=\frac{\partial}{\partial\beta}
E\Bigl[-\log Z_{\lambda}(\beta|\xi(u))\Bigr].
\]
It is conjectured that this value is determined by
the generalized Fisher information matrix $E_{X}[a(X,u)a(X,v)]$
on the set of true parameters ${\cal M}_{0}$. 
To investigate this problem in a mathematically rigorous way 
is a problem for  future study. 

Fifthly, we assumed that the log density ration function $f(x,w)$ is an $L^{s}(q)$-valued
analytic function. 
Even if $f(x, )$ is not analytic, if $f(x,)=u^{k}a(x,)$ holds and $a(x,)$ satisfies
some assuptions proved in Lemmas, then the theorem holds. 
However, if $f(x, )$ is not analytic, then 
there is examples in which $f(x, )=u^{k}a(x, )$ does not hold and it is not easy to judge
whether $f(x, )=u^{k}a(x, )$ holds or not. It is the future study the equations of states
in this paper in the more weak conditions. 

Lastly, let us compare the result of this paper
with the asymptotic theory of regular statistical models. 
In regular statistical models, the set of true parameters
consists of just one point, $W_{0}=\{w_{0}\}$. By the transform
$w=g_{0}(u)=w_{0}+I(w_{0})^{1/2}u$, where $I(w)$ is the
Fisher information matrix, 
\begin{eqnarray*}
K(g_{0}(u))&\cong&\frac{1}{2}|u|^{2},\\
K_{n}(g_{0}(u))&\cong&\frac{1}{2}|u|^{2}
-\frac{\xi_{n}}{\sqrt{n}}\cdot u,
\end{eqnarray*}
where $I(w_{0})$ is Fisher information matrix and 
$\xi_{n}=(\xi_{n}(1),\xi_{n}(2),...,\xi_{n}(d))$ is defined by 
\[
\xi_{n}(k)=\frac{1}{\sqrt{n}}\sum_{i=1}^{n}
\frac{\partial}{\partial u_{k}}
\log p(X_{i}|g_{0}(u))\Bigl{|}_{u=0}.
\]
Here each $\xi_{n}(k)$ converges in law
to the standard normal distribution.
Statistical learning theory for regular models is
based on the convergence in law $\xi_{n}\rightarrow \xi$,
whereas that for singular models, it is baesd on the fact that 
$\xi_{n}(u)\rightarrow \xi(u)$.

\section{Conclusion}

Based on singular learning theory, we established 
the equations of states in learning, and proposed
widely applicable information criteria.

\subsection*{Acknowledgment}
This research was partially supported by the Ministry of Education,
 Science, Sports and Culture in Japan, Grant-in-Aid for Scientific
 Research 18079007. 

\section{Appendix}\label{Chapter:Lemma}

\subsection{Proof of Lemma \ref{Lemma:1}}
Since $B_{g}(a)$ is the Kullback-Leibler distance from $q(x)$ to
$E_{w}[p(x|w)|_{K(w)\leq \epsilon}]$, 
$B_{g}(a)\geq 0$. Using Jensen's inequality, 
\[
E_{w}[e^{-f(x,w)}|_{K(w)\leq a}]\geq e^{-E_{w}[f(x,w)|_{K(w)\leq a}]}\;\;\;(\forall x),
\]
we have $B_{g}(a)\leq G_{g}(a)$ and $B_{t}(a)\leq G_{t}(a)$.
If $0<K(w)\leq a$, 
\begin{eqnarray*}
K_{n}(w) &= & K(w)-\sqrt{K(w)}\;\eta_{n}(w)\\
&\geq & (\sqrt{K(w)}-\frac{\eta_{n}(w)}{2})^{2}
-\frac{\eta_{n}(w)^{2}}{4}\\
&\geq & - \frac{1}{4}H_{t}(a).
\end{eqnarray*}
Hence $-H_{t}(a)/4\leq G_{t}(a)$. Also we have
\begin{equation}
K_{n}(w)\leq \frac{3}{2}K(w)+\frac{1}{2}\eta_{n}(w)^{2}.
\label{eq:cauchy}
\end{equation}
Therefore $G_{t}(a)\leq \frac{3}{2}G_{g}(a)+ \frac{1}{2}H_{t}(a)$. (Q.E.D.)

\subsection{Proof of Lemma \ref{Lemma:2}}

(1) For any $\epsilon>0$ and $a>0$, by the definition of $\eta_{n}(w)$, 
\[
\sqrt{n}\;\eta_{n}(w)=\frac{1}{\sqrt{K(w)}}\cdot\frac{1}{\sqrt{n}}
\sum_{j=1}^{n}(E_{X}[f(X,w)]-f(X_{j},w))
\]
is an empirical process and $f(x,w)$ is an analytic function of $w$,
hence 
\[
E[\sup_{\epsilon<K(w)<a}|\sqrt{n}\eta_{n}|^{6}]<const.
\]
\cite{Wvan}\cite{2005}\cite{morikita2006}. It is proven in Lemma
\ref{Lemma:5} that
$E[(nH_{t}(\epsilon))^{3}]$ also satisfies the same inequality.
(2) Let the random variable $S$ be defined by
\[
S=\left\{
\begin{array}{cc}
1 & (\mbox{ if } n H_{t}>n^{\alpha})\\
0 & (\mbox{ otherwise})
\end{array}\right..
\]
Then $E[S]=Pr(nH_{t}>n^{\alpha})$ and 
\[
C_{H}=E[(nH_{t})^{3}]
\geq E[(nH_{t})^{3}\;S]
\geq E[S]\;n^{3\alpha},
\]
which completes the Lemma. (Q.E.D.)

\subsection{Proof of Lemma \ref{Lemma:3}}

We use the notation, 
\begin{eqnarray*}
S_{1}(f(w))&=&\int_{K(w)\geq \epsilon} f(w)\;e^{-n\beta K_{n}(w)}\;\varphi(w)dw,\\
S_{0}(f(w))&=&\int_{K(w)<\epsilon}f(w)\; e^{-n\beta K_{n}(w)}\;\varphi(w)dw.
\end{eqnarray*}
By using the inequality,
\[
\frac{1}{2}K(w)-\frac{1}{2}\eta_{n}(w)^{2}\leq 
K_{n}(w) \leq  \frac{3}{2}K(w)+\frac{1}{2}\eta_{n}(w)^{2},
\]
we have inequalities for arbitrary $f(w),g(w)>0$, 
\begin{eqnarray*}
S_{1}(f(w)) & \leq  & \;(\sup_{w}f(w) )\;e^{-n\beta\epsilon/2}
\exp(\frac{\beta}{2}nH_{t}),\\
S_{0}(g(w)) & \geq & c_{0}\;(\inf_{w}g(w))\;n^{-\lambda}
\exp(-\frac{\beta}{2}nH_{t}),
\end{eqnarray*}
where $(-\lambda)$ is the largest pole of $\zeta(z)$ and 
$c_{0}>0$ is a constant which satisfies the inequality \cite{2001a}
\[
\int_{K(w)<\epsilon} \exp(-\frac{3\beta n}{2}K(w))\varphi_{1}(w)dw
\geq \frac{c_{0}}{n^{\lambda}}.
\]
Hence 
\[
\frac{S_{1}(f(w))}
{S_{0}(g(w))}
\leq 
\frac{\sup_{w}f(w)}{\inf_{w}g(w)}\;s(n),
\]
where 
\[
s(n)=\frac{n^{\lambda}}{c_{0}}\;e^{-n\beta\epsilon/2+n\beta H_{t}}.
\]
Then
\[
|\log s(n)|\leq n\beta\epsilon/2 + n\beta H_{t}+ \lambda \log n + |\log c_{0}|.
\]
By using the function $M(x)\geq 0 $ used in eq.(\ref{eq:M(x)}), we define $M_{n}$ by 
\[
M_{n}\equiv \frac{1}{n}\sum_{j=1}^{n}M(X_{j}).
\]
Then 
\[
E[M_{n}^{3}]\leq E[(\sum (M(X_{j})/n)^{3})]
\leq E[(\sum M(X_{j})^{3}/n)]=E_{X}[M(X)^{3}]<\infty. 
\]
(1) Firstly, we study Bayes generalization error.
\begin{eqnarray*}
n (B_{g}-B_{g}(\epsilon)) & = & 
n E_{X}
[
-\log 
\frac{E_{w}[e^{-f(X,w)}]}
{E_{w}[e^{-f(X,w)}|_{K(w)\leq \epsilon}]} ]\\
&=&
n E_{X}
[
-\log (1+\frac{S_{1}(e^{-f(X,w)})}
{S_{0}(e^{-f(X,w)})})
+\log (1+\frac{S_{1}(1)}
{S_{0}(1)})].
\end{eqnarray*}
Therefore
\begin{eqnarray*}
n | B_{g}-B_{g}(\epsilon)| & \leq &
n E_{X}
[
\log (1+
\frac{S_{1}(e^{-f(X,w)})}
{S_{0}(e^{-f(X,w)})})
+\log (1+\frac{S_{1}(1)}
{S_{0}(1)})] \\
&\leq &
n E_{X}[\log(1+s(n)\;e^{2\sup_{w}|f(X,w)|})
+\log (1+s(n))] \\
&\leq &
n E_{X}[\log(1+s(n)\;e^{2M(X)})]
+n s(n).
\end{eqnarray*}
The second term converges to zero in probability because of 
Lemma \ref{Lemma:2}. Let $f_{1}(n)$ be the first term,
\[
f_{1}(n)= n E_{X}[\log(1+s(n)\;e^{2M(X)})].
\]
Let us define
\begin{equation}
\Theta_{1}(x)=\left\{
\begin{array}{cc}
1 & (2M(x)> n\beta \epsilon/4) \\
0 & (2M(x)\leq n\beta\epsilon/4)
\end{array}
\right..
\end{equation}
Then by using $\log(1+x)\leq x$ and 
$\log(1+e^{x})\leq |x|+1$, 
\begin{eqnarray*}
f_{1}(n)&= & 
n E_{X}[(1-\Theta_{1}(X))\log(1+s(n)\;e^{2M(X)})]\\
&&+n E_{X}[\Theta_{1}(X)\log(1+s(n)\;e^{2M(X)})]\\
&\leq & n s(n)\exp(n\beta\epsilon/4)\\
&&+ nE_{X}[\Theta_{1}(X)(2M(X)+|\log s(n)|+1)],
\end{eqnarray*}
which converges to zero in probability because,
from the inequality eq.(\ref{eq:M(x)}), 
\begin{eqnarray*}
E_{X}[\Theta_{1}(X)M(X)]&\leq & (\frac{4}{n\beta\epsilon})^{5}E[M(X)^{6}],\\
E_{X}[\Theta_{1}(X)]&\leq & (\frac{4}{n\beta\epsilon})^{6}E[M(X)^{6}].
\end{eqnarray*}
It follows that $n(B_{g}-B_{g}(\epsilon))\rightarrow 0$. 
Secondly, we prove the convegence in probability $n(B_{t}-B_{t}(\epsilon))\rightarrow 0$. 
\begin{eqnarray}
n | B_{t}-B_{t}(\epsilon)| & \leq &
\sum_{j=1}^{n}\{
\log(1+s(n)\;e^{2\sup|f(X_{j},w)|})
+\log (1+s(n))\}\nonumber \\
& \leq &  \sum_{j=1}^{n}\log(1+s(n)e^{2M(X_{j})})+
n \log(1+s(n))\equiv L_{n}
\label{eq:Bt-prob}
\end{eqnarray}
where eq.(\ref{eq:Bt-prob}) is the definition of $L_{n}$. 
To prove the convergence in probability $L_{n}\rightarrow 0$, it is sufficient to
prove convergence in mean $E[L_{n}]\rightarrow 0$. Let the random variable 
$\Theta_{2}$ be 
\begin{equation}
\Theta_{2}=\left\{
\begin{array}{cc}
1 & (n H_{t}> n\beta \epsilon/4)\\
0 & (nH_{t} \leq n\beta\epsilon/4)
\end{array}\right..
\end{equation}
Then
\begin{eqnarray*}
E[L_{n}]&=&E[L_{n}(1-\Theta_{2})]+E[L_{n}\Theta_{2}]\\
&\leq & n E_{X}[\log(1+(n^{\lambda}/c_{0})\;e^{2M(X)-n\beta\epsilon/4})]\\
&& + n^{\lambda+1}\exp(-n\beta\epsilon/4)/c_{0}\\
&& + E[\Theta_{2}n(2M_{n}+|\log s(n)|+1)] \\
&& + E[\Theta_{2}n(|\log s(n)|+1)]
\end{eqnarray*}
The first term goes to zero can be proved in the same way as $f_{1}(n)\rightarrow 0$.
The second term goes to zero as a real sequence. 
Both the third and fourth terms go to zero because 
\begin{eqnarray*}
E[\Theta_{2}n M_{n}]&\leq & 
n Pr(nH_{t}>n)^{1/2}E[M_{n}^{2}]^{1/2},\\
E[n \Theta_{2}(n\beta\epsilon)]&= & n^{2}\beta\epsilon\;Pr(nH_{t}>n\beta\epsilon/4), \\
E[n \Theta_{2}(nH_{t})]&\leq  & 
n Pr(nH_{t}>n\beta\epsilon/4)^{1/2}E[(nH_{t})^{2}]^{1/2},
\end{eqnarray*}
and by using Lemma 2. 
Thus we obtain 
$n(B_{t}-B_{t}(\epsilon))\rightarrow 0$. 
Thirdly, the Gibbs generalization error 
can be estimated as
\begin{eqnarray}
n|G_{g}-G_{g}(\epsilon)|
& \leq & \Bigl{|}n\frac{S_{0}(K(w))+S_{1}(K(w))}{S_{0}(1)+S_{1}(1)}
-\frac{nS_{0}(K(w))}{S_{0}(1)}\Bigr{|}\nonumber \\
& \leq & 
\frac{nS_{1}(K(w))}{S_{0}(1)}+
\frac{nS_{0}(K(w))S_{1}(1)}{S_{0}(1)^{2}}\nonumber\\
&\leq & 2n \;\overline{K}\;s(n),\label{eq:Gg-prob}
\end{eqnarray}
which converges to zero in probability. Lastly, in the same way, the 
Gibbs training error satisfies
\begin{eqnarray*}
n|G_{t}-G_{t}(\epsilon)| & \leq  & 2n \;s(n)\;\sup_{w}|K_{n}(w)|\\
&\leq & 2n\;s(n)\;M_{n}
\end{eqnarray*}
which converges to zero in probability. \\
(2) Firstly, from Lemma \ref{Lemma:2}, $nH_{t}$ is AUI. 
Secondly, let us prove $nB_{t}$ is AUI. Let $L_{n}$ be the 
term in eq.(\ref{eq:Bt-prob}). Then 
\[
|nB_{t}|\leq |nB_{t}(\epsilon)|+L_{n}.
\]
Moreover, by employing a function, 
\[
b(s)=-\frac{1}{n}\sum_{j=1}^{n}\log E_{w}[e^{-sf(X_{j},w)}],
\]
there exists $0<s^{*}<1$ such that 
\[
nB_{t}=nb(1) = \sum_{j=1}^{n}
\frac{E_{w}[f(X_{j},w)e^{-s^{*}f(X_{j},w)}]}
{E_{w}[e^{-s^{*}f(X_{j},w)}]}.
\]
Hence 
\[
|nB_{t}|\leq \sum_{j=1}^{n}\sup_{w}|f(X_{j},w)|
\leq n M_{n}
\]
Therefore
\[
|nB_{t}|\leq |nB_{t}(\epsilon)|+
B^{*},
\]
where 
\[
B^{*} \equiv 
\left\{
\begin{array}{cc}
n M_{n}& (nH_{t}>\epsilon\beta n/4)\\
L_{n}
&(nH_{t}\leq \epsilon\beta n /4)
\end{array}\right..
\]
By summing the above equations, 
\[
E[|nB_{t}|^{3/2}]\leq E[2|nB_{t}(\epsilon)|^{3/2}]
+ E[2(B^{*})^{3/2}].
\]
In Lemma \ref{Lemma:5}, we prove that $E[|nB_{t}(\epsilon)|^{3/2}]<\infty$. 
By Lemma \ref{Lemma:2} (2) with $\delta$ such that 
$n^{\delta}=\epsilon\beta n /4$,
we have $P(H_{t}>\epsilon\beta/4)\leq C'_{H}/n^{3}$,
hence
\begin{eqnarray*}
E[(B^{*})^{3/2}]
& \leq & 
E[\Theta_{2}(B^{*})^{3/2}]
+
E[(1-\Theta_{2})(B^{*})^{3/2}] \\
&\leq & E[(nM_{n})^{3}]^{1/2}E[\Theta_{2}]^{1/2}\\
&&+ E[(1-\Theta_{2})(L_{n})^{3/2} ]<\infty.
\end{eqnarray*}
The first term is finite because $E[\Theta_{2}]=Pr(nH_{t}>n\beta\epsilon/4)$. 
Finiteness of the second term can be proved 
in the same way as proving that $E[(1-\Theta_{2})L_{n}]\rightarrow 0$. 
Hence $|nB_{t}|$ is AUI. 
Lastly, we show that $nG_{g}$ is AUI.
From eq.(\ref{eq:Gg-prob}),
\[
0\leq nG_{g}\leq  n G_{g}(\epsilon)+ 2n\;s(n)\;\overline{K}.
\]
Moreover, always $nG_{g}\leq n\overline{K}$, 
by definition. Therefore
\[
nG_{g}\leq nG_{g}(\epsilon)+ K^{*}
\]
where 
\begin{eqnarray*}
K^{*} & \equiv &
\left\{
\begin{array}{cc}
n\overline{K} & (nH_{t}>n^{2/3})\\
\overline{K}\;n\;s(n)
&(nH_{t}\leq n^{2/3})\\
\end{array}\right.\\
& \leq  &
\left\{
\begin{array}{cc}
n\overline{K} & (nH_{t}>n^{2/3})\\
\overline{K}\;e^{-n\beta\epsilon/3}
&(nH_{t}\leq n^{2/3})
\end{array}\right..
\end{eqnarray*}
Then 
\[
0\leq E[(nG_{g})^{3/2}]\leq 
E[2(nG_{g}(\epsilon))^{3/2}] + 
E[2(K^{*})^{3/2}].
\]
It is proven in Lemma \ref{Lemma:6} that
$
E[(nG_{g}(\epsilon))^{3/2}]<\infty.
$
By Lemma \ref{Lemma:2}  with $\delta=2/3$,
we have $P(nH_{t}>n^{2/3})\leq C_{H}/n^{2}$,
hence
\[
E[(K^{*})^{3/2}]
\leq n^{3/2}\overline{K}^{3/2}
\frac{C_{H}}{n^{2}}
+\overline{K}e^{-n\beta\epsilon/2}<\infty.
\]
Hence
$n G_{g}$ is AUI. Since $E[(nH_{t})^{3}]<\infty$, 
$E[(nB_{t})^{3/2}]<\infty$, and $E[(nG_{g})^{3/2}]<\infty$ 
all four errors are also AUI by Lemma \ref{Lemma:1}.
(Q.E.D.)

\subsection{Proof of Lemma \ref{Lemma:4}}

By the definition of the Kullback-Leibler distance and
 $f(x,g(u))=\log (q(x)/p(x|g(u)))$, 
for arbitrary $u\in{\cal M}$, 
\begin{eqnarray*}
K(g(u))&=&\int f(x,g(u))q(x)dx \\
&=&\int(e^{-f(x,g(u))}+f(x,g(u)))-1)q(x)dx\\
&=&\int \frac{f(x,g(u))^{2}}{2}e^{-t^{*}f(x,g(u))}q(x)dx,
\end{eqnarray*}
where $0<t^{*}<1$. 
Let $U'$ be a neighborhood of $u=0$. For arbitrary $L>0$ 
the set $D_{L}$ is defined by 
\[
D_{L}\equiv \{x\in {\bf R}^{N};\sup_{u\in U'}|f(x,g(u))|\leq L\}.
\]
Then for any $u\in U'$, 
\[
u^{2k}\geq \int_{D_{L}}\frac{f(x,g(u))^{2}}{2}e^{-L}q(x)dx,
\]
with the result that, for any $u^{k}\neq 0$ ($u\in U'$),
\begin{equation}\label{eq:smaller-one}
1\geq e^{-L}\int_{D_{L}}\frac{f(x,g(u))^{2}}{2u^{2k}}q(x)dx.
\end{equation}
Since $f(x,g(u))$ is an $L^{s}(q)$-valued real analytic function,
it is given by an absolutely convergent power series, 
\begin{eqnarray*}
f(x,g(u))&=& \sum_{\alpha}a_{\alpha}(x)u^{\alpha}\\
&=& a(x,u)u^{k}+b(x,u)u^{k},
\end{eqnarray*}
where 
\begin{eqnarray*}
a(x,u)&=&\sum_{\alpha\geq k}a_{\alpha}(x)u^{\alpha-k},\\
b(x,u)&=&\sum_{\alpha< k}a_{\alpha}(x)u^{\alpha-k},
\end{eqnarray*}
and $\sum_{\alpha\geq k}$ denotes the sum over indices that satisfy
\begin{equation}\label{eq:alpha-i}
\alpha_{i}\geq k_{i}\;\;\;(i=1,2,...,d)
\end{equation}
and $\sum_{\alpha<k}$ denotes the sum over indeces that do
not satisfy eq.(\ref{eq:alpha-i}). 
Here $a(x,u)$ is an $L^{s}(q)$-valued real analytic function.
From eq.(\ref{eq:smaller-one}), 
for an arbitrary $u^{k}\neq 0$ ($u\in U'$),
\begin{eqnarray*}
1 & \geq &  e^{-L}\int_{D_{L}}(a(x,u)+b(x,u))^{2}q(x)dx \\
&\geq &
\frac{e^{-L}}{2}\int_{D_{L}}b(x,u)^{2}q(x)dx
- e^{-L}\int_{D_{L}}a(x,u)^{2}q(x)dx.
\end{eqnarray*}
Here $|a(x,u)|$ is a bounded function of $u\in U'$.
If $b(x,u)\equiv 0$ does not hold, then 
$|b(x,u)|\rightarrow \infty$ $(u\rightarrow 0)$,
hence we can choose $u$ and $D_{L}$ 
so that the above inequality does not hold. Therefore, 
we have $b(x,u)\equiv 0$, which shows eq.(\ref{eq:Lemma4-1}). From 
\[
u^{2k}=\int f(x,g(u))q(x)dx=\int a(x,u)u^{k}q(x)dx,
\]
we obtain eq.(\ref{eq:Lemma4-2}). To prove eq.(\ref{eq:Lemma4-3}),
it is sufficient to prove $E_{X}[a(X,u)^{2}]=2$
when $K(g(u))=0$. 
Let the Taylor expansion of $f(x,g(u))$ be 
\[
f(x,g(u))=\sum_{\alpha}a_{\alpha}(x)u^{\alpha}.
\]
Then
\begin{equation}
|a_{\alpha}(x)|\leq \frac{M(x)}{R^{\alpha}}\label{eq:M(x)-R}
\end{equation}
where $R$ is the associated convergence radii 
and 
\[
a(x,u)=\sum_{\alpha\geq k}a_{\alpha}(x)u^{\alpha-k}.
\]
Hence 
\begin{eqnarray*}
|a(x,u)| & \leq & \sum_{\alpha\geq k}\frac{M(x)}{R^{\alpha}}r^{\alpha-k} \\
&=&c_{1}\frac{M(x)}{R^{k}},
\end{eqnarray*}
where $c_{1}>0$ is a constant. For arbitrary $u$ ($u^{k}\neq 0$), 
\[
1=\int\frac{a(x,u)^{2}}{2}e^{-t^{*}a(x,u)u^{k}}q(x)dx,
\]
where $0<t^{*}<1$. Put
\[
S(x,u)=\frac{a(x,u)^{2}}{2}e^{-t^{*}a(x,u)u^{k}}q(x).
\]
Then
\begin{eqnarray*}
S(x,u)&\leq & c_{1}\frac{M(x)^{2}}{R^{2k}}
\max\{1,e^{-a(x,u)u^{k}}\}q(x) \\
&=& 
c_{1}\frac{M(x)^{2}}{R^{2k}}
\max\{q(x),p(x|w)\}\\
&\leq & c_{1}\frac{M(x)^{2}}{R^{2k}}Q(x).
\end{eqnarray*}
By the fundamental condition (A.3), $M(x)^{2}Q(x)$ is 
an integrable function, hence $S(x,u)$ is bounded by 
the integrable function. 
By using Lebesgue's convergence theorem,
as $u^{k}\rightarrow 0$, we obtain
\[
1=\int \frac{a(x,u)^{2}}{2}q(x)dx
\]
for any $u$ that satisfies $u^{2k}=0$, which proves
eq.(\ref{eq:Lemma4-3}). Lastly, since $f(x,u)$ is an $L^{s}(q)$ valued
analytic function, $a(x,u)$ is also an $L^{s}(q)$ valued analytic function.
Moreover, eq.(\ref{eq:M(x)-R}) shows eq.(\ref{eq:Lemma4-4}).
(Q.E.D.)

\subsection{Proof of Lemma \ref{Lemma:5}}

The proof is given in \cite{2005} and
Theorem 39 in \cite{morikita2006}.

\subsection{Proof of Lemma \ref{Lemma:6}}


Let $u=(u_{1},u_{2},...,u_{d})$. Since at least one of non-negative integers $k_{1},..,k_{d}$ 
is not equal to zero, 
we can assume $k_{1}\geq 1$ without loss of generality. 
Put $g(u)=u_{2}^{k_{2}}\cdots u_{d}^{k_{d}}$ and $h(u)=u_{2}^{h_{2}}\cdots u_{d}^{k_{d}}$.
Then $u^{k}=u_{1}^{k_{1}}g(u)$, $u^{h}=u_{1}^{h_{1}}h(u)$,
where either $g(u)$ or $h(u)$ do not depend on $u_{1}$. 
We adopt the notation, 
\begin{eqnarray*}
N_{p}&=& \sum_{\alpha}\int_{[0,1]^{d}} u_{1}^{pk_{1}+h_{1}}g(u)^{p}h(u)\;
e^{-\beta n u^{2k}+f(u)}du, \\
f(u) &=& \beta\sqrt{n}u^{k}\xi(u)+\sigma u^{k}a(X,u),
\end{eqnarray*}
By the definition and $c_{1}=\|\psi\|/\|1/\psi\|$, 
\[
E_{u}^{\sigma}[u^{2k}|\xi]
\leq c_{1}\frac{N_{2}}{N_{0}}.
\]
By applying partial integration to $N_{2}$, 
\begin{eqnarray*}
N_{2} & = & 
-\sum_{\alpha}\int_{[0,1]^{d}} \frac{h(u)}{2\beta n k_{1}}\;
u_{1}^{h_{1}+1}e^{f(u)}\;
\partial_{1}(e^{-\beta n u^{2k}})\;du \\
&\leq & 
\sum_{\alpha}\int_{[0,1]^{d}} \frac{h(u)}{2\beta n k_{1}}\;
\partial_{1}(u_{1}^{h_{1}+1}e^{f(u)})\;
e^{-\beta n u^{2k}}\;du \\
&= & 
\sum_{\alpha}\int_{[0,1]^{d}} \frac{u_{1}^{h_{1}}h(u)}{2\beta n k_{1}}\;
e^{-\beta n u^{2k}+f(u)}\;
(h_{1}+1+u_{1}\partial_{1} f(u))\;du.
\end{eqnarray*}
From the definition of $f(u)$ 
\begin{eqnarray*}
u_{1}\partial_{1} f(u)
& = &\beta\sqrt{n}(k_{1}u^{k}\xi(u)+u^{k}\partial_{1} \xi(u))\\
&&+\sigma k_{1}u^{k}a(X,u)+\sigma u^{k}\partial_{1} a(X,u).
\end{eqnarray*}
By using inequalities
\begin{eqnarray*}
|\sqrt{n}u^{k}\xi(u) | & \leq & \frac{1}{2}(nu^{2k}+\xi(u)^{2}), \\
|\sqrt{n}u^{k}\partial_{1}\xi(u)|&\leq &
\frac{1}{2}(nu^{2k}+(\partial_{1}\xi(u))^{2}),
\end{eqnarray*}
and $|u^{k}|\leq 1$, 
\[
|u_{1}\partial_{1}f(u)|\leq 
\frac{\beta}{2}
\{k_{1}(nu^{2k}+\|\xi\|^{2})
+nu^{2k}+\|\partial_{1}\xi\|^{2}\}
+k_{1}\sigma\|a\|+\sigma\|\partial_{1} a\|.
\]
Hence
\[
\frac{N_{2}}{N_{0}}\leq 
\frac{1}{2n k_{1}}
\Bigl\{
\frac{n(k_{1}+1)}{2}
\frac{N_{2}}{N_{0}}
+h_{1}+1+
k_{1}\|\xi\|^{2}
+ \|\partial_{1}\xi\|^{2}
+ \frac{k_{1}\sigma\|a\|+\sigma \|\partial_{1}a\|}{\beta}
\Bigr\},
\]
with the result that 
\[
z_{1} \frac{N_{2}}{N_{0}}\leq 
\frac{1}{2n k_{1}}
\Bigl\{
h_{1}+1+
k_{1}\|\xi\|^{2}
+ \|\partial_{1}\xi\|^{2}
+ \frac{k_{1}\sigma\|a\|+\sigma \|\partial_{1}a\|}{\beta}
\Bigr\},
\]
where $z_{1}=(3k_{1}-1)/(4k_{1})$, 
which shows the first half of the lemma. Let us prove the latter half.
Firstly, 
\[
E_{u}^{\sigma}[u^{3k}|\xi]
\leq c_{3}\frac{N_{3}}{N_{0}}.
\]
In the same way as for the first half, by applying partial integration,
we have
\[
N_{3}\leq 
\sum_{\alpha}\int_{[0,1]^{d}} \frac{u^{h}u^{k}}{2\beta n k_{1}}\;
e^{-\beta n u^{2k}+f(u)}\;
(h_{1}+k_{1}+1+u_{1}\partial_{1} f(u))\;du.
\]
Therefore, we obtain
\begin{eqnarray*}
\frac{N_{3}}{N_{0}}
&\leq &
\frac{1}{2\beta k_{1} n}
\Bigl\{
\frac{N_{3}}{N_{0}}
\frac{n\beta(k_{1}+1)}{2}
+ \frac{N_{1}}{N_{0}}\\
& & \times ( k_{1}+h_{1}+1+
\frac{\beta k_{1}}{2}\|\xi\|^{2}
+\frac{\beta}{2}\|\partial_{1}\xi\|^{2}
+ k_{1}\sigma\|a\|+\sigma\|\partial_{1}a\|)
\Bigr\}.
\end{eqnarray*}
Therefore 
\[
z_{1}\frac{N_{3}}{N_{0}}\leq 
\frac{1}{2\beta k_{1} n}
\frac{N_{1}}{N_{0}}( k_{1}+h_{1}+1+
\frac{\beta k_{1}}{2}\|\xi\|^{2}
+\frac{\beta}{2}\|\partial_{1}\xi\|^{2}
+ k_{1}\sigma\|a\|+\sigma\|\partial_{1}a\|).
\]
By using Caucy-Schwarz inequality, that is to say, $N_{1}/N_{0}\leq (N_{2}/N_{0})^{1/2}$, and 
and by applying the result of the first half and H\"{o}lder's inequality,
\begin{eqnarray*}
\frac{N_{3}}{N_{0}}
& \leq & 
\frac{1}{2\beta k_{1} n}
(\frac{N_{2}}{N_{0}})^{1/2}
\Bigl\{k_{1}+h_{1}+1+
\frac{\beta k_{1}}{2}\|\xi\|^{2}
+\frac{\beta}{2}\|\partial_{1}\xi\|^{2}
+ k_{1}\sigma\|a\|+\sigma\|\partial_{1}a\|)
\Bigr\}\\
&\leq & \frac{C}{n^{3/2}}
\{1+\|\xi\|^{2}+\|\partial_{1}\xi\|^{2}+
\sigma\|a\|+\sigma\|\partial_{1}a\|\}^{3/2},
\end{eqnarray*}
where $C>0$ is a constant which is determined by $k_{1},h_{1}$, and $\beta$. 
In general, 
\[
(\frac{1}{5}\sum_{k=1}^{5}|a_{k}|^{2})^{3/2}\leq \frac{1}{5}\sum_{k=1}^{5} |a_{k}|^{3},
\]
which completes the proof. 
(Q.E.D.)

\subsection{Proof of Lemma \ref{Lemma:7}}

For given functions $\xi(u)$ and $g(u)$, we define 
\begin{eqnarray}
A^{p}(\xi,g)& \equiv &\sum_{\alpha}\int_{[0,1]^{r}}dx
\int_{[0,1]^{r'}}dy\;
(x^{2k}y^{2k'})^{p}\;x^{h}y^{h'}\;g(x,y)\nonumber \\
&&\times e^{-n\beta x^{2k}y^{2k'}+\sqrt{n}\beta x^{k}y^{k'}\xi(x,y)}.
\label{eq:ap}
\end{eqnarray}
Then 
\[
E_{u}^{0}[u^{2pk}f(u)|\xi]
=\frac{A^{p}(\xi,f\psi)}{A^{0}(\xi,\psi)}.
\]
It is rewritten as
\[
A^{p}(\xi,g)=\sum_{\alpha}
\int_{0}^{\infty}dt
\int dx\; dy \;\delta(t-nx^{2k}y^{2k'})\;
x^{h}y^{h'}g(x,y)\;\frac{t^{p}}{n^{p}}\;e^{-\beta t-\beta \sqrt{t}\xi(x,y)}.
\]
To analyze $\delta(\cdot)$ function, we need the fact that, for $Re(z)>0$, 
\begin{eqnarray*}
\int_{[0,1]^{r}} (a\;x^{2k})^{z}\;x^{h}\;dx
&=&a^{z} \prod_{j=1}^{r}\int_{0}^{1}x_{j}^{2k_{j}z+h_{j}}\;dx_{j}\\
&=&\frac{a^{z}}{2^{r}\;k_{1}\cdots k_{r}\;(z+\lambda_{\alpha})^{r}}.
\end{eqnarray*}
By applying the inverse Mellin transform to this equation, 
we have
\[
\int_{[0,1]^{r}}
\delta(t-ax^{2k})\;x^{h}\;dx
=\left\{
\begin{array}{cc}
c_{0}\frac{t^{\lambda_{\alpha}-1}}{a^{\lambda_{\alpha}}}
(\log\frac{a}{t})^{r-1} &(0<t<a) \\
0 &(\mbox{otherwise})
\end{array}\right.
\]
where $c_{0}=1/(2^{r}(r-1)!k_{1}\cdots k_{r})$. If
$g_{0}(y)=g(0,y)$ and $\xi_{0}(y)=\xi(0,y)$ then
\begin{eqnarray}
A^{p}(\xi_{0},g_{0})
= \sum_{\alpha}\int_{0}^{\infty}dt \int_{t<ny^{2k'}<n}dy\;
c_{0}\frac{y^{\mu}t^{p+\lambda_{\alpha}-1}}
{n^{p+\lambda_{\alpha}}}  e^{-\beta t-\beta \sqrt{t}\xi_{0}(y)}
(\log\frac{ny^{2k'}}{t})^{r-1}g_{0}(y).\label{eq:xy-txy}
\end{eqnarray}
where the region `$t<ny^{2k'}<n$' denotes the set $\{y\in [0,1]^{s};t<ny^{2k'}<n\}$.
Then by using eq.(\ref{eq:xy-txy}),
\begin{eqnarray}
|A^{p}(\xi,g)| 
& \leq & c_{0} \|g\|e^{-\beta \|\xi\|^{2}/2}
\int_{0}^{\infty}dt \int_{[0,1]^{r}}dy\;
\frac{y^{\mu}t^{p+\lambda_{\alpha}-1}}{n^{p+\lambda_{\alpha}}}
|\log\frac{ny^{2k'}}{t}|^{r-1}\;e^{-\beta t/2}\nonumber \\
&\leq & c_{1}\|g\|e^{-\beta \|\xi\|^{2}/2}\frac{(\log n)^{r-1}}{n^{p+\lambda_{\alpha}}},
\label{eq:up-bd}
\end{eqnarray}
where $c_{1}>0$ is a constant. In the same way, 
\begin{equation}
|A^{p}(\xi,g)|\geq 
c_{1}'\min|g|\;e^{-3\beta\|\xi\|^{2}/2}\frac{(\log n)^{r-1}}{n^{p+\lambda_{\alpha}}}.
\label{eq:low-bd}
\end{equation}
Let $\lambda$ be the smallerst value in $\{\lambda_{\alpha};\alpha\}$. Then
$(-\lambda)$ is equal to the largest pole of $\zeta(z)$. 
The coordinate $U_{\alpha}$ whose $\lambda_{\alpha}$ is
equal to the smallest one $\lambda_{\alpha}=\lambda$ and whose 
$r$ is equal to the largest one $r=m$
is denoted by $U_{\alpha^{*}}$. The sum $\sum_{\alpha^{*}}$ 
denotes the sum restricted to such coordinates.
Let $A_{*}^{p}(\xi,g)$ be the sum of $A^{p}(\xi,g)$ restricted in this way,
in other words, $\sum_{\alpha}$ is replaced by $\sum_{\alpha^{*}}$ in
eq.(\ref{eq:ap}).
Also we define 
$C^{p}_{*}(\xi,g)=A^{p}_{*}(\xi,g)-A^{p}_{*}(\xi_{0},g_{0})$. 
There exists $x^{*}\in [0,1]^{r}$ such that
\[
e^{-\beta\sqrt{t}\xi(x,y)}g(x,y)-e^{-\beta\sqrt{t}\xi(0,y)}g(0,y)
=
\sum_{j=1}^{r}x_{j}
\{\partial_{j}g(x^{*},y)-\beta\sqrt{t}g\partial_{j}\xi(x^{*},y)\}
e^{-\beta \sqrt{t}\xi(x^{*},y)}
\]
Hence
\begin{equation}
|C^{p}_{*}(\xi,g)|\leq 
c_{2}(\|\nabla g\| + \beta \|g\| \|\nabla \xi\| )\;
e^{-\beta\|\xi\|^{2}/2}\frac{(\log n)^{m-2}}{n^{p+\lambda}}.
\label{eq:Remain-bd}
\end{equation}
By expanding eq.(\ref{eq:xy-txy}), we have
\begin{eqnarray*}
A^{p}_{*}(\xi_{0},g_{0})& = &\sum_{k=1}^{m}A^{pk}_{*}(\xi_{0},g_{0}) \\
A^{pk}_{*}
(\xi_{0},g_{0}) &=& \sum_{\alpha^{*}}
\frac{(\log n)^{k-1}}{n^{p+\lambda}}{m-1 \choose k-1}
\int_{0}^{\infty}dt \int_{t<ny^{2k'}<n}dy \\
& & \times c_{0} \; y^{\mu}\;
(\log \frac{y^{2k'}}{t})^{m-k}g_{0}(y)\;
t^{p+\lambda-1}\;e^{-t+\sqrt{t}\xi_{0}(y)}.
\end{eqnarray*}
The largest order term among them is $A^{pm}_{*}(\xi_{0},g_{0})$.
We define $B^{pm}_{*}(\xi_{0},g_{0}$ from  $A^{pm}_{*}$ by
replacing the integral region of $y$, 
\begin{eqnarray*}
B^{pm}_{*}(\xi_{0},g_{0}) &=& \sum_{\alpha^{*}}
\frac{(\log n)^{m-1}}{n^{p+\lambda}}
\int_{0}^{\infty}dt \int_{[0,1]^{r}}dy \\
& & \times c_{0} \; y^{\mu}\;
g_{0}(y)\;t^{p+\lambda-1}\;e^{-\beta t+\beta \sqrt{t}\xi_{0}(y)}.
\end{eqnarray*}
The difference between $A^{pm}_{*}(\xi_{0},g_{0})$ and 
$B^{pm}_{*}(\xi_{0},g_{0})$ 
is smaller than $\|g\|e^{-\|\xi\|^{2}/2}/n^{p+\lambda}$, and 
\begin{eqnarray}
|A^{pk}_{*}(\xi_{0},g_{0})| & \leq & c_{3}\|g\|e^{-\beta\|\xi\|^{2}/2}
\;\frac{(\log n)^{k-1}}{n^{p+\lambda}}
\;\;\;(1\leq k\leq m),\label{eq:zzz1} \\
|B^{pm}_{*}(\xi_{0},g_{0})| & \geq & c_{3'}\|g\|e^{-3\beta\|\xi\|^{2}/2}
\;\frac{(\log n)^{r-1}}{n^{p+\lambda}}. \label{eq:zzz2}
\end{eqnarray}
By the definition, 
\[
D\equiv E_{u}^{0}[u^{2pk}f(u)|\xi]-E_{y,t}[t^{p}f(0,y)|\xi]
= \frac{A^{p}(\xi,f\psi)}{A^{0}(\xi,\psi)} -
\frac{
B^{pm}_{*}(\xi_{0},f_{0}\psi_{0})}{
B^{0m}_{*}(\xi_{0},\psi_{0})}.
\]
Then using eqs.(\ref{eq:up-bd})-(\ref{eq:zzz2}),
\begin{eqnarray*}
R^{p}(\xi,g) &\equiv & A^{p}(\xi,g)-B^{pm}_{*}(\xi_{0},g_{0}) \\
&=& A^{p}_{o}(\xi,g) 
+ C^{p}_{*}(\xi,g)+\sum_{k=1}^{m}A^{pk}_{*}(\xi_{0},g_{0})
- B^{pm}_{*}(\xi_{0},g_{0}),
\end{eqnarray*}
where $A^{p}_{o}(\xi,g)=A^{p}(\xi,g)-A^{p}_{*}(\xi,g)$ is the sum
over $\alpha$ that are not $\alpha^{*}$. 
Therefore
\[
|R^{p}(\xi,g)|\leq \frac{c_{4}}{n^{p+\lambda}}\;e^{-\beta\|\xi\|^{2}}\;
(\|g\|+\beta\|g\|\|\nabla \xi\|+\|\nabla g\|)
\]
Thus
\begin{eqnarray*}
|D| & \leq & \frac{|R^{p}(\xi,f\psi)|}{A^{0}(\xi,\psi)}
+ 
\frac{|R^{0}(\xi,\psi)||B^{pm}_{*}(\xi_{0},f_{0}\psi_{0})|}
{A^{0}(\xi,\psi)B^{0m}_{*}(\xi_{0},\psi_{0})}\\
& \leq & \|\psi\|\|\frac{1}{\psi}\|
\frac{c_{5}}{n^{p}\log n} \;e^{4\beta\|\xi\|^{2}}\;
(\|g\|+\beta\|g\|\|\nabla \xi\|+\|\nabla g\|)
\end{eqnarray*}
which completes the Lemma. (Q.E.D.)

\subsection{Proof of Lemma \ref{Lemma:8}}

By using partial integration, 
for an arbitrary $a\in R$, 
\[
\int_{0}^{\infty}
e^{-\beta t}\;
2t^{\lambda}\;e^{\beta a \sqrt{t}}
\;dt
=
\frac{1}{\beta}\int_{0}^{\infty}e^{-\beta t}
\frac{\partial}{\partial t}\Bigl(
2t^{\lambda}\;e^{\beta a \sqrt{t}}\Bigr)\;dt.
\]
Hence
\begin{equation}
\int_{0}^{\infty}dt\;
(2t-\sqrt{t}a-\frac{2\lambda}{\beta})\;
t^{\lambda-1}\;e^{-\beta t+\beta\sqrt{t}a}=0.
\label{eq:2t-ta-lambda}
\end{equation}
which shows Lemma \ref{Lemma:8}.
(Q.E.D.)

\end{document}